\documentclass[a4paper,11pt]{article}
\usepackage[margin=1in]{geometry}

\usepackage{amsmath,amsfonts,bm}

\def\eqref#1{equation~\ref{#1}}

\def\1{\bm{1}}

\DeclareMathAlphabet{\mathsfit}{\encodingdefault}{\sfdefault}{m}{sl}
\SetMathAlphabet{\mathsfit}{bold}{\encodingdefault}{\sfdefault}{bx}{n}

\usepackage{amssymb}
\usepackage{graphicx}
\usepackage{natbib}
\usepackage{booktabs}
\usepackage{xcolor}
\usepackage{url}
\usepackage{parskip}
\usepackage{hyperref}
\hypersetup{
    colorlinks,
    linkcolor={blue!90!black},
    citecolor={magenta},
    urlcolor={blue!80!black}
}

\usepackage{cleveref}
\usepackage{multirow}
\usepackage{float}
\usepackage{array}

\title{Advancing Mass-Conserving Deep Learning for Distributed Hydrologic Modeling: A Case Study in a Snow-Dominated Watershed}

\author{Yihan Wang$^{1}$, \,\, Lujun Zhang$^{1}$, \,\, N. Benjamin Erichson$^{2,3}$, \,\, Yang Hong$^{1}$, \\ and Tiantian Yang$^{1,4}$\thanks{Corresponding author: \url{yangtt@umich.edu}} \vspace{+0.3cm} \\
\parbox{\textwidth}{\centering\small
$^1$ School of Civil Engineering and Environmental Science, University of Oklahoma, Norman, OK, United States \\
$^2$ Lawrence Berkeley National Laboratory, Berkeley, CA, United States \\
$^3$ International Computer Science Institute, Berkeley, CA, United States \\
$^4$ School for Environment and Sustainability, University of Michigan, Ann Arbor, United States}
}
\date{}

\begin{document}

\maketitle


\begin{abstract}

Recent progress in physics-aware deep learning (DL) has embedded mass conservation within Long Short-Term Memory (LSTM) architecture for hydrologic modeling, improving physical consistency and enabling diagnostics of internal states. However, existing mass-conserving neural networks are primarily developed in lumped formulations. Consequently, previous studies provide only fragmented and often indirect evidence of when and why imposed physical constraints may lead to unrealistic hydrologic behavior. To address this limitation, we propose Physics-Enhanced Conservation-Augmented Net (PECAN), a mass-conserving convolutional LSTM-based framework designed not only for distributed hydrologic modeling but also for diagnosing structural limitations of physics-aware DL models. PECAN is implemented in two variants, namely (1) PECAN\textsubscript{MC} with a strict precipitation-based conservation constraint and (2) PECAN\textsubscript{MCR} with a relaxed conservation constraint by allowing additional water inputs beyond precipitation. Both variants are~trained solely on gauge streamflow and evaluated in a snow-dominated watershed in Colorado. We benchmark PECAN against lumped DL baselines and the distributed process-based Variable Infiltration Capacity (VIC) model. Results show that PECAN achieves competitive outlet streamflow simulation accuracy while producing spatially explicit latent states for interpretation. Comparisons with reference soil moisture indicate that PECAN cell states resemble meaningful temporal soil moisture dynamics. However, analyses also reveal that PECAN misrepresents snow processes by prematurely removing winter snow water storage from the system rather than retaining it for spring release. In summary, PECAN demonstrates accurate streamflow simulation at gauges and shows diagnostic potential to expose structural limitations in hydrologic process representation, thereby informing future physics-aware neural network development for hydrology.

\end{abstract}

\section{Introduction}
\label{sec:introduction}

Deep Learning (DL) models have demonstrated strong predictive accuracy in hydrologic modeling. However, their opaque internal mechanisms and insufficient adherence to physical principles hinder understanding of when and why DL models succeed or fail, raising concerns regarding interpretability, generalization to unseen conditions, and scientific credibility. These concerns have motivated growing interest in physics-aware DL models, which incorporate domain knowledge into data-driven models to reconcile the predictive power with physical consistency \citep{reichstein2019deep,vonrueden2021informed}.

Physics-aware DL models can be broadly categorized into soft-constraint and hard-constraint approaches, based on how strictly physical principles are enforced \citep{wang2025investigating}. Soft-constraint approaches embed physical knowledge indirectly, such as by augmenting loss functions to penalize violations of physical laws. While straightforward to implement, these strategies do not guarantee physical consistency during inference and thus provide limited reliability \citep{beucler2021enforcing,karniadakis2021physics}. In contrast, hard-constraint approaches enforce physical principles through the model's mathematical or architectural design, imposing a structural prior on the behaviors the network can represent, and thereby offering a stricter alternative to soft-constraint methods.

Among hard-constraint formulations for hydrologic modeling, mass conservation, a fundamental property of hydrologic systems, has received particular interest. \citet{hoedt2021mc} pioneered the explicit incorporation of a precipitation-based mass conservation constraint into the Long Short-Term Memory (LSTM; \citealp{hochreiter1997long}) architecture, introducing the Mass Conserving (MC) LSTM. In MC-LSTM, the cell state is interpreted as a conceptual water storage, and the gating functions are reformulated such that water inputs, outputs, and storage changes satisfy mass balance by design. Subsequent studies evaluated the hydrologic capability of MC-LSTM for large-scale hydrologic modeling \citep{frame2023strictly,frame2022deep,wi2024need}. \citet{wang2025mcr} further introduced the Mass Conservation Relaxed (MCR) LSTM, which relaxes the precipitation-only conservation assumption in MC-LSTM to allow for additional unobserved water inputs and to reflect more realistic hydrologic conditions. Beyond sequence-level constraints applied within LSTM architectures, mass conservation has also been enforced at the individual perceptron level \citep{wang2025towards,wang2024mass,wang2024towards,wang2025using}. Together, these developments demonstrate that hard-constrained mass conservation can substantially influence model performance and potentially bring new hydrologic insights by enabling the interpretation of DL model internal state and flux dynamics.

Despite this progress, hard-constrained mass-conserving neural networks have been focused almost exclusively on lumped hydrologic modeling, where conservation principles are applied to basin-aggregated water fluxes and storage states. The lumped formulations are generally most suitable for small to moderate catchments or applications focused on outlet-scale prediction, where spatial heterogeneity can be reasonably averaged \citep{bloschl1995scale}. However, this basin-level aggregation becomes limiting in larger or more heterogeneous domains, where hydrologic behavior depends on spatial variability in meteorological forcing and runoff generation across the landscape \citep{tran2018spatially}. More importantly, such a collapse of spatial variability also constrains scientific understanding of internal model behavior. Specifically, existing lumped mass-conserving DL models restrict diagnostics to basin-averaged time series \citep[e.g.,][]{wang2025mcr}, which are insufficient to verify whether the learned latent states encode physically meaningful within-basin storage patterns and flow partitioning. Consequently, our understanding of how conservation constraints interact with data-driven learning remains incomplete, highlighting a need to extend hard-constrained physics-aware DL models toward distributed hydrologic formulations.

Recent efforts to address this need are largely built on the differentiable parameter learning (dPL; \citealp{feng2022differentiable,shen2023differentiable,tsai2021calibration}) framework. In dPL, a process-based hydrologic model serves as the computational backbone, and neural networks are used to learn model parameters in an end-to-end differentiable manner. Hydrologic states are updated according to predefined process equations, and physical constraints such as mass conservation are inherently enforced by the mathematical structure of the underlying hydrologic model \citep{he2025distributed,wang2024distributed,zhong2024advancing,zhong2024development}. The dPL framework has demonstrated considerable promise for combining physical consistency with the flexibility of deep learning in distributed hydrologic modeling. At the same time, it also highlights an opportunity to examine the role of neural network architecture itself in physics-aware DL. Specifically, instead of achieving physical fidelity through a process-based model backbone, a contrasting direction is to impose only a minimal set of fundamental physical constraints (e.g., mass conservation) in a DL model, thereby enabling investigation of how such constraints shape information propagation within a neural network and whether they can support any degree of autonomous hydrologic discovery.

To further address the gap in the current physics-aware DL landscape, we investigate the direct incorporation of mass conservation constraints within a distributed neural network architecture. We aim to address the following research questions concerning model design, interpretability, and the hydrologic effects of mass-conservation assumptions: (1) How can a mass-conserving neural architecture be developed for spatially distributed hydrologic modeling? (2) Are the learned distributed internal states and fluxes hydrologically meaningful and interpretable? (3) How do different mass-conservation assumptions shape the model's learned hydrologic behavior?

To answer these questions, we propose the Physics-Enhanced Conservation-Augmented Net (PECAN), a convolutional (Conv) LSTM \citep{shi2015convolutional}-based framework that imposes mass-conservation constraints directly within the neural architecture for distributed hydrologic modeling. PECAN is formulated in two variants: PECAN\textsubscript{MC}, which enforces a strict precipitation-based constraint motivated by \citet{hoedt2021mc}, and PECAN\textsubscript{MCR}, which relaxes this constraint by allowing additional water inputs beyond precipitation, motivated by \citet{wang2025mcr}. Compared with the lumped mass-conserving counterparts, PECAN introduces several structural adaptations that enable end-to-end distributed hydrologic modeling, including spatially resolved water-storage states and runoff generation, a two-storage structure motivated by rapid and delayed hydrologic responses, and physically based routing of grid-cell runoff to downstream gauges.

As a proof of concept, both PECAN variants are trained and evaluated at the daily scale in the snow-dominated North Platte Headwaters watershed in Colorado. Model performance is benchmarked against a suite of lumped DL baselines, including a standard LSTM as well as MC-LSTM and MCR-LSTM, which serve as lumped mass-conserving counterparts to PECAN. In addition, the Variable Infiltration Capacity (VIC; \citealp{liang1994simple}) is included as a distributed process-based hydrologic baseline. Beyond modeling accuracy, we assess the physical interpretability of PECAN's internal states by comparing their spatiotemporal behavior with independent hydrologic reference datasets, including observation-based snow water equivalent (SWE) and soil moisture, as well as VIC-simulated soil moisture and evapotranspiration (ET). Finally, PECAN\textsubscript{MC} and PECAN\textsubscript{MCR} are intercompared to investigate how strict versus relaxed mass conservation constraints influence the model's representation of hydrologic processes.

\section{Dataset and Study Site}
\label{sec:dataset_study_site}

\subsection{Dataset}
\label{sec:dataset}

The input meteorological forcings for PECAN include daily precipitation, maximum temperature, and minimum temperature, which are obtained from the Parameter-elevation Regressions on Independent Slopes Model (PRISM; \citealp{daly2013prism}). PRISM is an observation-based climate dataset with a 4-km spatial resolution over the contiguous United States (CONUS). It employs digital elevation models (DEM) and weighted regressions to account for the effects of elevation, topography, coastal proximity, and orographic processes. Due to its ability to resolve spatial variability in complex terrain, PRISM has been frequently adopted as a meteorological forcing dataset for hydrologic and land-surface modeling, particularly in mountainous and snow-dominated regions \citep{iseri2021dynamical,rudisill2024cold,shuai2021effects,yue2025machine,zhang2024adapting,zhang2025alternative}.

The observation-based SWE and soil moisture datasets used for PECAN internal state interpretation are obtained from the University of Arizona SWE (UA-SWE) dataset \citep{broxton2019daily} and from the Soil Moisture Active Passive (SMAP) Level-4 (L4) soil moisture product \citep{reichle2017assessment}, respectively. The UA-SWE dataset is generated through a data assimilation framework that integrates snow model simulations with in situ SWE observations, including SNOTEL measurements. The resulting dataset provides gridded, spatially continuous SWE estimates over CONUS at a 4-km spatial resolution and daily temporal resolution. The SMAP L4 product provides surface (0--5 cm) and root zone (0--100 cm) soil moisture estimates at approximately 9-km spatial resolution and 3-hourly temporal resolution by assimilating SMAP brightness temperature observations into the NASA Catchment Land Surface Model. The original SMAP spatial resolution is preserved without interpolation to the PECAN grid for spatial visual comparison. To ensure temporal consistency with PECAN outputs, the SMAP L4 soil moisture fields are aggregated to daily means.

\subsection{Study Site}
\label{sec:study_site}

\begin{figure}[!htbp]
\centering
\includegraphics[width=\textwidth]{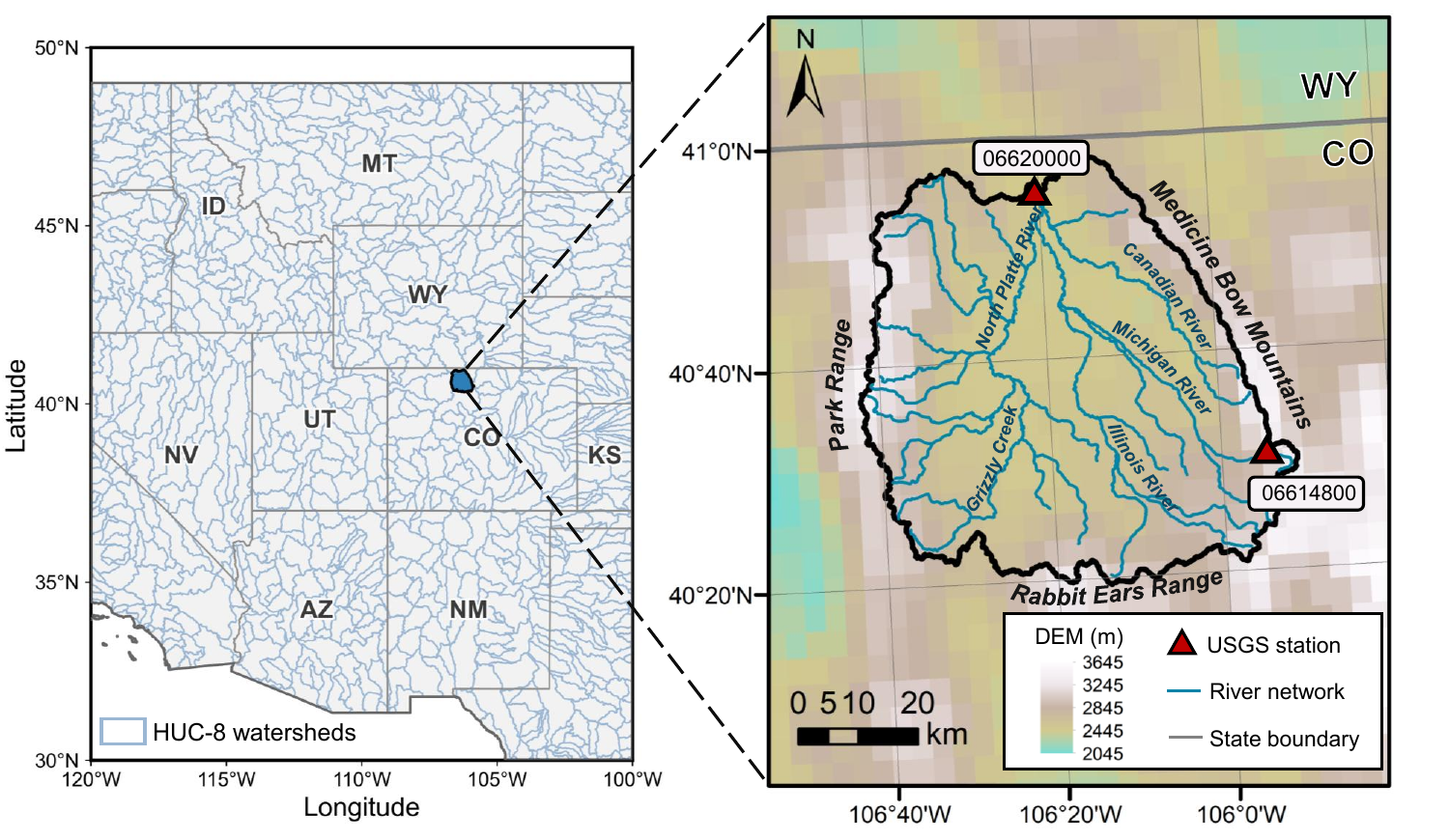}
\vspace{-0.7cm}
\caption{Maps of the North Platte Headwaters watershed. The left panel shows the watershed location within the western United States. The right panel shows the watershed elevation, boundary, and river network. Red triangles indicate USGS streamflow stations used in this study.}
\label{fig:study_site}
\end{figure}

The study site is the North Platte Headwaters, a Hydrologic Unit Code level 8 (HUC-8) watershed located in the north-central Colorado within the upper Colorado-Wyoming Rocky Mountains, as shown in Figure~\ref{fig:study_site}. The watershed drainage area is approximately 3,688 km$^2$ of mountainous terrain. Basin elevations range from approximately 2,381 m at the outlet to nearly 3,945 m along the surrounding ridgelines that define the watershed boundary. The basin is enclosed by prominent mountain ranges, including the Medicine Bow Mountains to the east, the Rabbit Ears Range to the south, and the Park Range to the west. Major tributaries within the basin include the Michigan River, Illinois River, Grizzly Creek, and Canadian River, which drain high-elevation snow-accumulation zones before joining the main stem of the North Platte River. Streamflow observations are obtained from two USGS stream gauges: station 06620000 (North Platte River near Northgate, CO), which serves as the basin outlet, and station 06614800 (Michigan River Near Cameron Pass, CO), located upstream within the headwater network.

The watershed exhibits a cold, snow-dominated hydroclimate typical of high-elevation Rocky Mountain basins. A large fraction of annual precipitation occurs during winter and early spring and is stored seasonally as snowpack. Precipitation exhibits strong elevation gradients, ranging from roughly 300 mm/year in valley bottoms to more than 1000 mm/year along high-elevation ridges \citep{daniels2007flood}. Hydrologic behavior is characterized by strong seasonal storage--release dynamics. Streamflow typically shows low winter baseflow followed by a pronounced snowmelt-driven rise in late spring and early summer, with peak flows commonly occurring during May to June, and recession flows through late summer. Vegetation and land cover vary strongly with elevation. High elevation zones are dominated by subalpine conifer forests and alpine tundra, while mid- to low-elevation valleys are covered by grasslands and riparian corridors along the river network \citep{acharya2011modeled}. Downstream of the headwaters, the North Platte River flows through a system of large reservoirs in Wyoming that regulate water supply and hydroelectric production, making seasonal inflows from the headwaters critical for regional water management \citep{daniels2007flood}.

\section{Methodology}
\label{sec:methodology}

In this study, we develop and evaluate two distributed PECAN variants, PECAN\textsubscript{MC} and PECAN\textsubscript{MCR}, and compare them against three lumped deep learning baselines (LSTM, MC-LSTM, and MCR-LSTM) and a distributed process-based VIC baseline. Section~\ref{sec:pecan} introduces the overall architecture of PECAN. Section~\ref{sec:baseline_models} describes the baseline models. Section~\ref{sec:deep_learning_model} presents the training configurations for PECAN and the baseline deep learning models. Section~\ref{sec:evaluation} describes the evaluation metrics and performance assessment strategy used for model comparison.

\subsection{PECAN}
\label{sec:pecan}

\begin{figure}[!htbp]
\centering
\includegraphics[width=\textwidth]{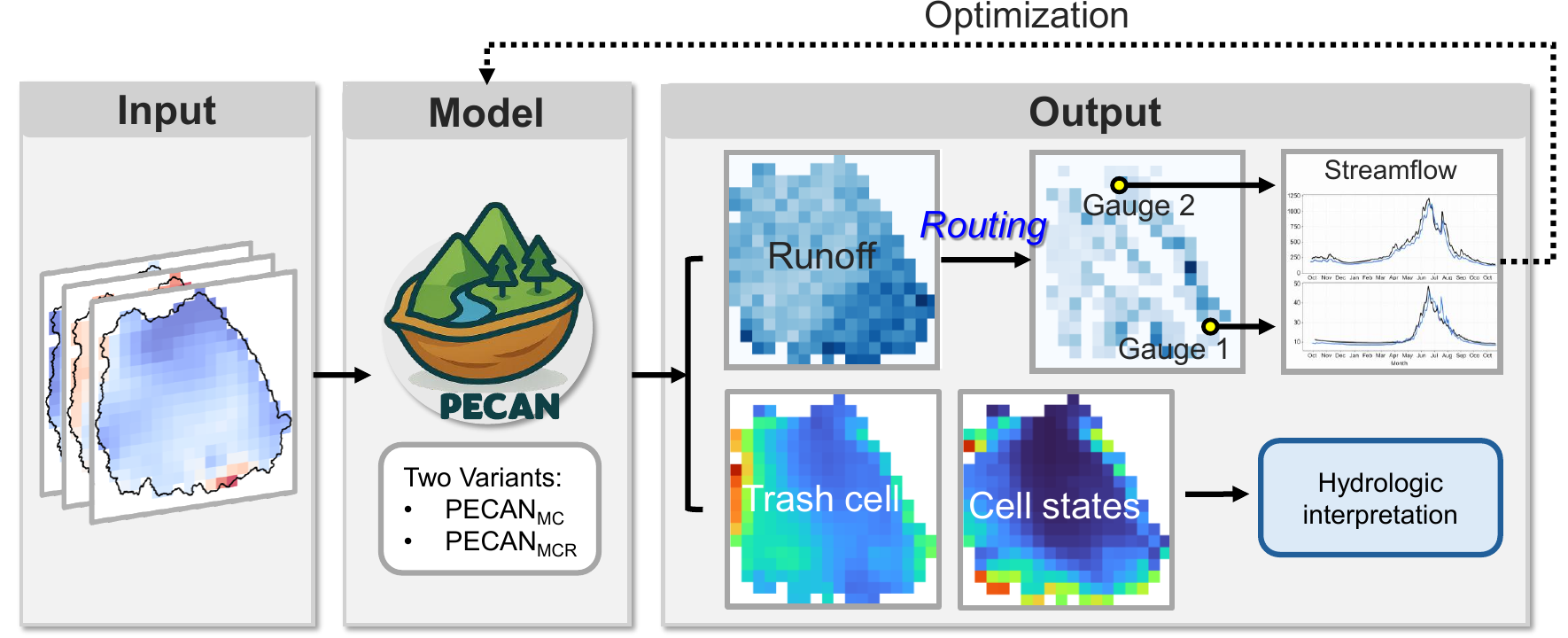}
\vspace{-0.7cm}
\caption{Overall workflow of the Physics-Enhanced Conservation-Augmented Network (PECAN). Gridded meteorological forcings are used as spatial inputs to PECAN, which is implemented in two variants (i.e., PECAN\textsubscript{MC} and PECAN\textsubscript{MCR}). The model produces spatially distributed runoff, which is routed to gauge locations to generate streamflow predictions for optimization against streamflow observations. PECAN also produces trash cell flux and cell states for hydrologic interpretation.}
\label{fig:workflow}
\end{figure}

PECAN extends the ConvLSTM architecture by embedding explicit water balance constraints within its gating mechanisms while preserving spatial dynamics through convolutional operators. It consists of two tightly coupled components: a mass-conserving runoff module and a physically based routing module, described in Sections~\ref{sec:runoff_module} and~\ref{sec:routing_module}, respectively. An overview of PECAN's workflow, inputs, and outputs is presented in Figure~\ref{fig:workflow}.

PECAN is implemented in two conservation formulations, PECAN\textsubscript{MC} and PECAN\textsubscript{MCR}. Both variants share the same two-layer storage structure, reflecting a common hydrologic separation between a fast-response component associated with quick runoff generation and a slow-response component associated with delayed water release. At each time step, gridded precipitation enters the model as the mass input, while other meteorological variables such as temperature are treated as auxiliary inputs that help regulate the partitioning of water without directly contributing mass. The incoming water is then distributed among two storage layers, released flux, and water loss removal at each grid cell, thereby preserving an explicit water accounting. The two PECAN variants differ in how strictly this accounting is constrained by precipitation as the external water source. Specifically, in PECAN\textsubscript{MC}, precipitation is treated as the only external water source, such that changes in storage and water release/loss must be explained entirely by precipitation together with previously stored water. In PECAN\textsubscript{MCR}, the precipitation-based conservation is relaxed through a bidirectional mass relaxation pathway, allowing the model to account for unresolved inputs beyond precipitation and underrepresented losses. Together, the two variants provide a framework for examining how strict versus relaxed mass conservation influences both simulation performance and learned internal behavior.

In a standard LSTM, the cell state acts as an internal memory that retains information over time, while the hidden state carries the model output from one time step to the next. In PECAN, these two states are given explicit hydrologic interpretations. The cell states represent spatially distributed fast- and slow-flow water storage, whereas the hidden states represent the corresponding fast- and slow-flow water fluxes released from these storages. For the fast-flow component, the first hidden channel is designated as the trash cell flux. In the original MC-LSTM design, the trash cell discards excess mass that does not contribute to the prediction, which in hydrologic applications has been interpreted as an aggregate of unresolved sinks, including ET \citep{hoedt2021mc,frame2022deep,wang2025mcr}. The remaining hidden channels represent fast runoff generation. Similarly, the slow-flow contribution represents delayed runoff generation. Together, the fast and slow runoff components constitute the total runoff, which is subsequently propagated through the routing module to generate streamflow at gauge locations. Model training is performed exclusively using observed discharge at gauges extracted from the routed streamflow, without supervision from other distributed target variables. The internal states, including the storage-like cell state and the trash cell flux, are retained for hydrologic interpretation and diagnostic comparison with independent observations and VIC-simulated hydrologic states.

\subsubsection{Runoff Module}
\label{sec:runoff_module}

\begin{figure}[!t]
\centering
\includegraphics[width=\textwidth]{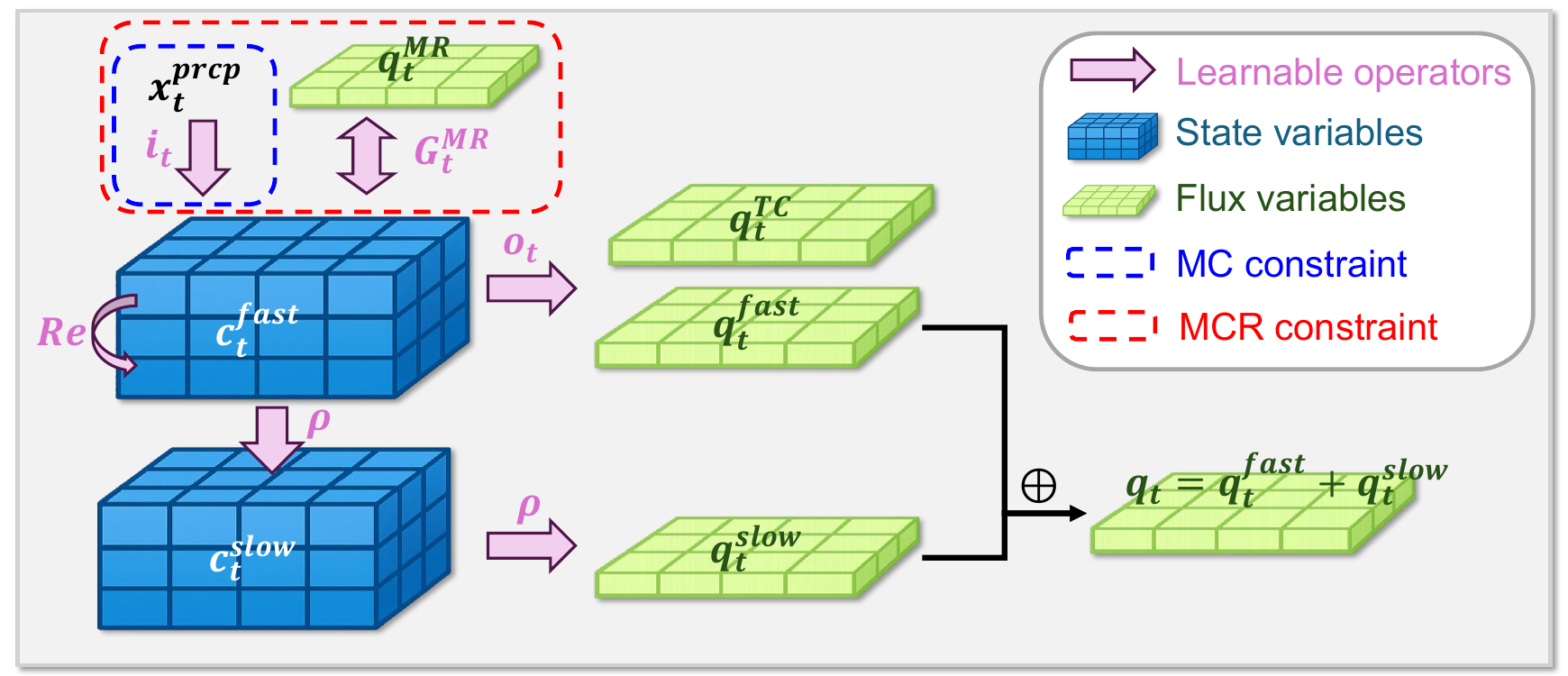}
\vspace{-0.7cm}
\caption{Schematic illustration of the PECAN runoff module. The fast-flow storage $\mathbf{c}_{t}^{\mathrm{fast}}$ and slow-flow storage $\mathbf{c}_{t}^{\mathrm{slow}}$ evolve through conservation-constrained gating mechanisms. Under mass conservation (MC) constraint, the input gate $\mathbf{i}_{t}$ partitions precipitation $\mathbf{x}_{t}^{\mathrm{prcp}}$ into fast-flow storage channels, the redistribution operator $\mathbf{Re}$ mixes existing fast-flow storage, and the output gate $\mathbf{o}_{t}$ controls the release flux from the fast-flow storage. This released flux consists of the trash cell flux $\mathbf{q}_{t}^{\mathrm{TC}}$ representing water sinks such as ET or unresolved losses, as well as the fast runoff $\mathbf{q}_{t}^{\mathrm{fast}}$. Under the mass conservation relaxed (MCR) constraint, an additional bidirectional relaxation gate $\mathbf{G}_{t}^{\mathrm{MR}}$ is included to allow for additional mass injection or removal within the fast-flow storage. The leakage coefficient $\rho$ governs both recharge from fast-flow storage to slow-flow storage, as well as discharge from the slow-flow storage as slow runoff $\mathbf{q}_{t}^{\mathrm{slow}}$. The effective runoff field used for routing excludes the trash cell and sums the fast and slow runoff.}
\label{fig:runoff_module}
\end{figure}

PECAN represents the watershed as a spatially distributed two-level storage system consisting of a fast-flow storage and a slow-flow storage. At each time step, precipitation is partitioned into fast-flow storage channels through a convolutional input gate, redistributed internally, and then released as fast runoff through a convolutional output gate. A fraction of the fast-flow storage is transferred to the slow-flow storage, which produces delayed runoff through a linear reservoir formulation. The runoff module then computes total runoff as the sum of the fluxes released from both storages. A schematic of the runoff module is shown in Figure~\ref{fig:runoff_module}.

PECAN operates on a spatial grid of size $H\times W$, defined by the meteorological forcing data. At time $t$, PECAN takes precipitation input $\mathbf{x}_{t}^{\mathrm{prcp}}\in \mathbb{R}^{1\times H\times W}$ and auxiliary inputs $\mathbf{x}_{t}^{\mathrm{aux}}\in \mathbb{R}^{C_{a}\times H\times W}$, both masked by $\mathbf{M}\in \{0,1\}^{1\times H\times W}$ (with 1 indicating grid cells within the watershed) so that all hydrologic processes are restricted to the watershed grid cells. Grid cells outside the watershed ($M=0$) are treated as inactive, effectively imposing a no-flow boundary condition. The fast-flow storage and slow-flow storage states are defined as $\mathbf{c}_{t}^{\mathrm{fast}}, \mathbf{c}_{t}^{\mathrm{slow}}\in \mathbb{R}^{C\times H\times W}$, respectively, where $C$ is the number of cell state channels. The individual channels are latent dimensions that jointly represent each storage component and are not assigned separate hydrologic meanings. Both storage tensors share the same spatial resolution and mask as the forcing inputs.

\textbf{Mass Input}. PECAN conditions its gates on input forcings and a channel-wise L1-normalized fast-flow storage as given by Equation~(\ref{eq:norm_storage}), where $\epsilon$ is a small constant (1e-8) for numerical stability.
\begin{equation}
\tilde{\mathbf{c}}_{t-1}^{\mathrm{fast}} = 0.9\,\frac{\mathbf{c}_{t-1}^{\mathrm{fast}}}{\left\| \mathbf{c}_{t-1}^{\mathrm{fast}} \right\|_{1,\mathrm{ch}} + \epsilon} + 0.1\,\mathbf{c}_{t-1}^{\mathrm{fast}}
\label{eq:norm_storage}
\end{equation}

The unnormalized input gate is then computed as given in Equation~(\ref{eq:input_gate_raw}),
\begin{equation}
\hat{\mathbf{i}}_{t} = \sigma\left( \mathrm{GN}\left( \mathbf{W}_{i} \circledast \mathbf{x}_{t}^{\mathrm{prcp}} + \mathbf{U}_{i} \circledast \tilde{\mathbf{c}}_{t-1}^{\mathrm{fast}} + \mathbf{V}_{i} \circledast \mathbf{x}_{t}^{\mathrm{aux}} + \mathbf{b}_{i} \right) \right)
\label{eq:input_gate_raw}
\end{equation}
where $W_{i}$, $U_{i}$, and $V_{i}$ are learnable convolution kernels, $b_{i}$ is a bias term, $\circledast$ denotes two-dimensional convolution, $\mathrm{GN}(\cdot)$ denotes channel-wise group normalization (G=8) for training stability, and $\sigma(\cdot)$ is the sigmoid function. To ensure precipitation mass is conservatively partitioned across channels, the input gate is normalized along the channel dimension, as in Equation~(\ref{eq:input_gate}).
\begin{equation}
\mathbf{i}_{t} = \frac{\hat{\mathbf{i}}_{t}}{\sum_{k=1}^{C} \hat{\mathbf{i}}_{t}^{(k)} + \epsilon}
\label{eq:input_gate}
\end{equation}

The incoming mass to the fast-flow storage is computed as the element-wise (Hadamard) product of precipitation and the normalized input gate, as given in Equation~(\ref{eq:mass_in}).
\begin{equation}
\mathbf{m}_{t}^{\mathrm{in}} = \mathbf{x}_{t}^{\mathrm{prcp}} \odot \mathbf{i}_{t}
\label{eq:mass_in}
\end{equation}

\textbf{Mass Redistribution}. Existing fast flow storage is redistributed across channels through a learned redistribution matrix $\mathbf{Re}\in \mathbb{R}^{C\times C}$. $\mathbf{Re}$ is a nonnegative parameter matrix shared across grid cells, and it is column-normalized to ensure mass conservation at each grid cell, as given in Equation~(\ref{eq:redistribution}),
\begin{equation}
\mathbf{Re} = \frac{\mathrm{ReLU}(\boldsymbol{\Theta})}{\sum_{k=1}^{C} \mathrm{ReLU}(\boldsymbol{\Theta})_{k} + \epsilon}
\label{eq:redistribution}
\end{equation}
where $\boldsymbol{\Theta}\in \mathbb{R}^{C\times C}$ is a learnable parameter matrix. The redistribution acts only along the channel dimension within each grid cell and does not represent spatial water transport.

The total water mass before release is therefore the sum of the incoming precipitation mass and the redistributed fast-flow storage, as given in Equation~(\ref{eq:mass_new}).
\begin{equation}
\mathbf{m}_{t}^{\mathrm{new}} = \mathbf{m}_{t}^{\mathrm{in}} + \mathbf{Re}\,\mathbf{c}_{t-1}^{\mathrm{fast}}
\label{eq:mass_new}
\end{equation}

\textbf{Mass Output and Storage Update for MC and MCR}. The output gate $\mathbf{o}_{t}$ controls the released flux $\mathbf{q}_{t}^{\mathrm{rel}}$ from the total water mass, as in Equations~(\ref{eq:output_gate}) and~(\ref{eq:release}),
\begin{equation}
\mathbf{o}_{t} = \sigma\left( \mathrm{LN}_{\mathrm{ch}}\left( \mathbf{W}_{o} \circledast \mathbf{x}_{t}^{\mathrm{prcp}} + \mathbf{U}_{o} \circledast \tilde{\mathbf{c}}_{t-1}^{\mathrm{fast}} + \mathbf{V}_{o} \circledast \mathbf{x}_{t}^{\mathrm{aux}} + \mathbf{b}_{o} \right) \right)
\label{eq:output_gate}
\end{equation}
\begin{equation}
\mathbf{q}_{t}^{\mathrm{rel}} = \mathbf{o}_{t} \odot \mathbf{m}_{t}^{\mathrm{new}}
\label{eq:release}
\end{equation}
where $\mathrm{LN}_{\mathrm{ch}}(\cdot)$ denotes channel-wise layer normalization to prevent channel dominance. The first channel of $\mathbf{q}_{t}^{\mathrm{rel}}$ is designated as the trash cell flux $\mathbf{q}_{t}^{\mathrm{TC}}$, representing water sinks such as ET or unresolved losses. Fast runoff $\mathbf{q}_{t}^{\mathrm{fast}}$ is then given by Equation~(\ref{eq:fast_runoff}).
\begin{equation}
\mathbf{q}_{t}^{\mathrm{fast}} = \mathbf{q}_{t}^{\mathrm{rel}} - \mathbf{q}_{t}^{\mathrm{TC}}
\label{eq:fast_runoff}
\end{equation}

In the mass conserving formulation implemented in PECAN\textsubscript{MC}, the updated fast-flow storage is given by Equation~(\ref{eq:storage_mc}).
\begin{equation}
\mathbf{c}_{t}^{\mathrm{fast}} = (1 - \mathbf{o}_{t}) \odot \mathbf{m}_{t}^{\mathrm{new}}
\label{eq:storage_mc}
\end{equation}

In the relaxed mass conservation formulation implemented in PECAN\textsubscript{MCR}, an additional mass relaxation (MR) gate $\mathbf{G}_{t}^{\mathrm{MR}}$ is introduced to allow adaptive bidirectional mass exchange between the fast-flow storage and the flux component. The MR gate is constructed based on the formulation of MCR-LSTM in \citet{wang2025mcr}, but is adapted into a spatially distributed convolutional form, as shown in Equation~(\ref{eq:mr_gate}):
\begin{equation}
\mathbf{G}_{t}^{\mathrm{MR}} = \mathrm{LN}_{\mathrm{ch}}\left( \tanh\left( \mathrm{LN}_{\mathrm{ch}}\left( \left( \tilde{\mathbf{c}}_{t-1}^{\mathrm{fast}} - \mathbf{b}_{\mathrm{MR}} \right) \circledast \mathbf{W}_{\mathrm{MR}} \right) \right) \circledast \boldsymbol{\kappa}_{\mathrm{MR}} \right)
\label{eq:mr_gate}
\end{equation}
where $\mathbf{W}_{\mathrm{MR}}$ and $\boldsymbol{\kappa}_{\mathrm{MR}}$ are learnable convolution kernels, $\boldsymbol{\kappa}_{\mathrm{MR}}\in [0,1]$ controls the relaxation magnitude, and $b_{\mathrm{MR}}$ is a learnable bias. The hyperbolic tangent activation $\tanh(\cdot)$ ensures that the MR gate is inherently signed, allowing both positive and negative relaxation. Compared with the original MCR-LSTM formulation, PECAN additionally applies channel-wise LayerNorm operations before the tanh activation to prevent saturation under large storage magnitudes. Following \citet{wang2025mcr}, the MR gate is constrained elementwise per grid cell and per channel between $[-\mathbf{o}_{t},1-\mathbf{o}_{t}]$ to preserve the nonnegativity of the cell state. The relaxation flux $\mathbf{q}_{t}^{\mathrm{MR}}$ induced by the MR gate and the corresponding updated fast-flow storage are given by Equations~(\ref{eq:mr_flux}) and~(\ref{eq:storage_mcr}), respectively.
\begin{equation}
\mathbf{q}_{t}^{\mathrm{MR}} = \mathbf{G}_{t}^{\mathrm{MR}} \odot \mathbf{m}_{t}^{\mathrm{new}}
\label{eq:mr_flux}
\end{equation}
\begin{equation}
\mathbf{c}_{t}^{\mathrm{fast}} = (1 - \mathbf{o}_{t} - \mathbf{G}_{t}^{\mathrm{MR}}) \odot \mathbf{m}_{t}^{\mathrm{new}}
\label{eq:storage_mcr}
\end{equation}
Finally, the slow-flow storage $\mathbf{c}_{t}^{\mathrm{slow}}$ is represented as a first-order linear reservoir. A learnable fraction $\rho \in (0,1)$ of the fast-flow storage recharges the slow-flow storage, while discharge from the slow-flow storage is linearly proportional to its stored water. For simplicity, the same coefficient $\rho$ governs both recharge and discharge processes. The slow-flow dynamics are therefore given by Equations~(\ref{eq:slow_storage}) and~(\ref{eq:slow_runoff}). Lastly, the total runoff generated at each grid cell before routing is the sum of fast and slow runoff, as given in Equation~(\ref{eq:total_runoff}).
\begin{equation}
\mathbf{c}_{t}^{\mathrm{slow}} = (1 - \rho)\,\mathbf{c}_{t-1}^{\mathrm{slow}} + \rho\,\mathbf{c}_{t}^{\mathrm{fast}}
\label{eq:slow_storage}
\end{equation}
\begin{equation}
\mathbf{q}_{t}^{\mathrm{slow}} = \rho\,\mathbf{c}_{t}^{\mathrm{slow}}
\label{eq:slow_runoff}
\end{equation}
\begin{equation}
\mathbf{q}_{t} = \mathbf{q}_{t}^{\mathrm{fast}} + \mathbf{q}_{t}^{\mathrm{slow}}
\label{eq:total_runoff}
\end{equation}

For hydrologic interpretation and model output, the latent channels are aggregated at each grid cell. The fast- and slow-flow cell states are each summed across channels to obtain their respective spatial storage fields. Likewise, the released fluxes are summed across channels, with the trash cell channel excluded from runoff.

\subsubsection{Routing Module}
\label{sec:routing_module}

PECAN employs the RVIC routing model to convert gridded runoff into streamflow at gauge locations. RVIC is a source-to-sink routing model adapted from the VIC routing framework of \citet{lohmann1996large,lohmann1998regional}, and has been used as an offline post-processing routing scheme for VIC runoff at regional to continental scales \citep{chegwidden2019modeling,niemeyer2018thermally,zheng2026what}. For each source grid cell and target gauge, RVIC precomputes an impulse-response function, or unit hydrograph, that describes the timing and dispersion of runoff as it travels downstream. The flow path and travel distance are determined from the 4-km D8 flow-direction network, while transport along that path is represented using a linearized one-dimensional Saint-Venant formulation:
\begin{equation}
\frac{\partial Q}{\partial t} = D\frac{\partial^{2} Q}{\partial x^{2}} - C\frac{\partial Q}{\partial x}
\label{eq:saint_venant}
\end{equation}
where $Q$ is discharge, $x$ is downstream distance, $C$ is flow-wave velocity that controls the travel time of the runoff response, and $D$ is hydraulic diffusivity that controls its spreading and attenuation. Both $C$ and $D$ are parameters that can be prescribed from available geographic or hydraulic information, or calibrated. The corresponding impulse response over a routing distance $x$ is:
\begin{equation}
h(x,t) = \frac{x}{2t\sqrt{\pi D t}} \exp\left[ -\frac{(Ct-x)^{2}}{4Dt} \right]
\label{eq:impulse_response}
\end{equation}

For this study, spatially varying values of $C$ are derived from NHDPlus High Resolution mean-annual velocity estimates \citep{moore2025nhdplus}, using the value from the dominant channel occupying that cell. $D$ is prescribed uniformly as $800~\mathrm{m^{2}/s}$.

In addition to downstream channel routing shown in Equation~(\ref{eq:impulse_response}), RVIC represents the transfer of locally generated runoff within each source grid cell through a within-cell unit hydrograph $U(t)$. Here, $U(t)$ is prescribed uniformly as a normalized, short-duration response that peaks at 1 h and decays to zero after 5 h, representing sub-grid runoff concentration before water enters the downstream routing network. The complete source-to-gauge impulse response is then obtained by convolving this within-cell response with the downstream routing response $h(x, t)$:
\begin{equation}
H_{g,i}(t) = \int_{0}^{t} U(t-s)\, h_{g,i}(s)\, ds
\label{eq:source_to_gauge}
\end{equation}
where $H_{g,i}(t)$ is the final impulse-response function relating runoff generated at grid cell $i$ to streamflow at gauge $g$. The resulting impulse-response functions are generated before PECAN training and stored as a fixed routing kernel. Streamflow at gauge $g$ is then calculated as:
\begin{equation}
Q_{g}(t) = \sum_{i=1}^{N} \sum_{\tau=0}^{K-1} H_{g,i}(\tau)\, q_{i}(t-\tau)
\label{eq:gauge_flow}
\end{equation}
where $q_{i}(t)$ is runoff generated by PECAN at grid cell $i$, $N$ is the number of source grid cells, and $K$ is the routing-kernel length. During model training, the RVIC routing kernel remains fixed and is used solely to propagate runoff downstream, while only the runoff-generation module is optimized.

\subsection{Baseline Models}
\label{sec:baseline_models}

Three lumped DL models---LSTM, MC-LSTM, and MCR-LSTM---and the distributed process-based VIC 4.2.d model are included as baselines. The standard LSTM provides a widely used data-driven benchmark for rainfall--runoff modeling, while MC-LSTM and MCR-LSTM provide mass-conserving counterparts for comparison with PECAN\textsubscript{MC} and PECAN\textsubscript{MCR}, respectively. Each lumped architecture is evaluated under both local and regional training. The local training provides a direct basin-specific comparison with PECAN, whereas the regional training serves as a stronger large-sample benchmark that incorporates information shared across many watersheds. The corresponding training configurations are described in Section~\ref{sec:deep_learning_model}.

The VIC model is implemented on the same 4-km watershed grid as PECAN. Because the PRISM forcings used in this study are at daily resolution, VIC is run at a daily timestep in water-balance mode, with both the land-surface and snow calculations evaluated at 24-h intervals. Daily precipitation and temperature are obtained from PRISM, while wind speed is obtained from gridMET \citep{abatzoglou2013development}. Initial soil, vegetation, vegetation-library, and five-band snow-elevation parameters are obtained from the CONUS VIC parameter dataset by \citet{livneh2015spatially} and mapped from their native grid to the 4-km model grid using nearest-neighbor sampling. VIC parameters, including the ARNO baseflow parameters $D_{\mathrm{smax}}$, and $W_{s}$, the variable-infiltration parameter $b_{\mathrm{infilt}}$, total soil column depth, and minimum stomatal resistance $r_{\min}$ are calibrated using the Shuffled Complex Evolution algorithm (SCE-UA; \citealp{duan1994optimal}). The model is calibrated over WY1985--WY2012 and evaluated over WY2013--WY2022, with WY1982--WY1984 used as a spin-up period. The calibration objective is the mean of NSE and KGE at the outlet. VIC-generated runoff and baseflow are then routed offline using the same RVIC source-to-sink routing framework applied to PECAN, as described in Section~\ref{sec:routing_module}.

\subsection{Deep Learning Model Training}
\label{sec:deep_learning_model}

PECAN and the lumped LSTM, MC-LSTM, and MCR-LSTM baselines are trained and validated using 31 water years of daily data spanning WY 1982 to WY 2012, corresponding to the combined spin-up and calibration period used for VIC, with 85\% of the samples randomly assigned to training and 15\% for validation. The testing period spans from WY2013 to WY2022, consistent with the evaluation period used for VIC. Model inputs include daily precipitation, maximum air temperature (Tmax), and minimum air temperature (Tmin) from PRISM. For PECAN, the forcings are provided as spatially distributed gridded inputs across the watershed, whereas the lumped baselines use basin-averaged time series derived from the same gridded products. The standard LSTM baseline uses normalized inputs for all variables. For mass-conserving architectures (MC-LSTM, MCR-LSTM, PECAN\textsubscript{MC}, and PECAN\textsubscript{MCR}), precipitation is retained in physical units to preserve explicit mass accounting, while auxiliary temperature inputs are normalized.

PECAN\textsubscript{MC} and PECAN\textsubscript{MCR} are trained against two USGS gauges within the basin (06620000 at the outlet and 06614800 upstream) using a weighted Nash-Sutcliffe Efficiency (NSE; \citealp{nash1970river}). The outlet gauge is assigned a larger weight to prioritize basin-scale streamflow performance, while the upstream gauge provides an additional constraint on the spatial distribution of runoff generation. The gauge weights are treated as model hyperparameters rather than being prescribed according to drainage-area proportion. Following tuning, weights of 0.8 and 0.2 are assigned to the outlet and upstream gauges, respectively, for PECAN\textsubscript{MC}, while weights of 0.9 and 0.1 are used for PECAN\textsubscript{MCR}.

Two sets of lumped baselines are trained. The locally trained LSTM, MC-LSTM, and MCR-LSTM use the outlet gauge only with the standard NSE loss, and are tuned specifically for the study watershed. The regional baselines follow configurations established in previous CONUS-wide rainfall-runoff simulation studies \citep{kratzert2019towards,wang2025mcr} and are trained jointly across the 531 CAMELS watersheds and the study watershed, resulting in a total of 532 watersheds. The regional training uses the same PRISM meteorological forcings as locally trained baselines (precipitation, Tmax, and Tmin), together with 27 static catchment attributes widely used in previous CONUS-wide studies (Table S1 in the Appendix).

All models adopt a sequence-to-one training formulation. For each training sample, the cell states in PECAN (both fast-flow and slow-flow storage states) and lumped models are initialized to zero to avoid reliance on prescribed initial storage values and enable a generic application. A sufficiently long input sequence allows the models to construct an effective internal representation of antecedent hydrologic conditions. Only the runoff at the final time step of each sequence is used for supervision. All models are optimized using the Adam optimizer \citep{kingma2014adam}. The locally trained baselines and PECAN use a cosine annealing learning rate schedule \citep{loshchilov2016sgdr}, which initializes with a relatively large learning rate and smoothly decays it following a cosine trajectory to promote stable convergence. The regional baselines instead follow the model configurations adopted from the previous large-sample studies \citep{frame2023strictly,frame2022deep,wang2025mcr}. Hyperparameters for the local baselines and PECAN, including hidden dimension, sequence length, batch size, number of epochs, and initial learning rate, are manually tuned. All configurations are summarized in Table~\ref{tab:training_config}.

\begin{table}[!htbp]
\centering
\caption{Training configurations and hyperparameters for PECAN and the local and regional lumped deep learning baselines.}
\label{tab:training_config}
\vspace{+0.1cm}
\resizebox{\textwidth}{!}{%
\begin{tabular}{lccccc}
\toprule
\textbf{Model} & \textbf{Hidden channels} & \textbf{Sequence length} & \textbf{Epoch} & \textbf{Batch size} & \textbf{(Init.) Learning rate} \\
\midrule
\multicolumn{6}{c}{Lumped baselines (local)} \\
\midrule
LSTM     & 64 & 180 & 100 & \multirow{3}{*}{64} & \multirow{3}{*}{1e-3} \\
MC-LSTM  & 32 & 60  & 80  &  &  \\
MCR-LSTM & 32 & 60  & 50  &  &  \\
\midrule
\multicolumn{6}{c}{Lumped baselines (regional)} \\
\midrule
LSTM     & 256 & 365 & 30 & \multirow{3}{*}{1024} & \multirow{3}{*}{\parbox{4.2cm}{\centering 1e-3 for epochs 1-10, 5e-4 for epochs 11-25, 1e-4 for epochs 26-30}} \\
MC-LSTM  & 64  & 365 & 30 &  &  \\
MCR-LSTM & 64  & 365 & 30 &  &  \\
\midrule
\multicolumn{6}{c}{Distributed PECAN} \\
\midrule
PECAN\textsubscript{MC}  & 32 & 60  & 150 & \multirow{2}{*}{64} & 1.1e-3 \\
PECAN\textsubscript{MCR} & 48 & 180 & 200 &  & 8e-4 \\
\bottomrule
\end{tabular}%
}
\end{table}

\subsection{Evaluation}
\label{sec:evaluation}

Model evaluation follows a three-step design that combines (i) simulation performance assessment at streamflow gauges, (ii) internal model state interpretation using independent observation-based SWE and soil moisture, together with VIC-simulated ET and soil moisture, and (iii) diagnostics of how the mass conservation formulations (MC versus MCR) alter inferred water sources and sinks. The lumped baseline models, including LSTM, MC-LSTM, and MCR-LSTM, are used only as streamflow performance benchmarks. Subsequent process-level diagnostics focus on PECAN and VIC because their distributed structures permit spatially explicit analysis of states and fluxes.

First, the performance of PECAN and lumped baseline models is evaluated using NSE, Kling--Gupta Efficiency (KGE; \citealp{gupta2009decomposition}), the Pearson correlation coefficient (COR), percent bias in flow duration curve high-flow segment (top 2\%) volume (FHV; \citealp{yilmaz2008process}), percent bias in the low-flow segment (bottom 30\%) of the flow duration curve (FLV; \citealp{yilmaz2008process}), and the overall percentage bias (PBias). In addition to these statistical summaries, hydrographs for selected water years are presented for PECAN\textsubscript{MC} and PECAN\textsubscript{MCR} at both gauges, with VIC simulations included at the outlet gauge only, to visualize differences in peak magnitude, timing, and recession behavior that are not fully captured by scalar metrics. Metric formulations are provided in Equations~(\ref{eq:nse}) to~(\ref{eq:pbias}), where $n$ denotes the length of the testing time series, $H$ is the number of high-flow observations with exceedance probability below 0.02, and $L$ is the number of low-flow observations within the 0.7--1.0 exceedance-probability range.
\begin{equation}
\mathrm{NSE} = 1 - \frac{\sum_{i=1}^{n} (Q_{\mathrm{obs},i} - Q_{\mathrm{sim}})^{2}}{\sum_{i=1}^{n} (Q_{\mathrm{obs},i} - \overline{Q}_{\mathrm{obs}})^{2}}
\label{eq:nse}
\end{equation}
\begin{equation}
\mathrm{KGE} = 1 - \sqrt{(CC-1)^{2} + \left( \frac{\overline{Q}_{\mathrm{sim}}}{\overline{Q}_{\mathrm{obs}}} - 1 \right)^{2} + \left( \frac{\sigma_{\mathrm{sim}}}{\sigma_{\mathrm{obs}}} - 1 \right)^{2}}
\label{eq:kge}
\end{equation}
\begin{equation}
\mathrm{COR} = \frac{\sum_{i=1}^{n} \left( (Q_{\mathrm{sim},i} - \overline{Q}_{\mathrm{sim}})(Q_{\mathrm{obs},i} - \overline{Q}_{\mathrm{obs}}) \right)}{\sqrt{\sum_{i=1}^{n} (Q_{\mathrm{sim},i} - \overline{Q}_{\mathrm{sim}})^{2} \sum_{i=1}^{n} (Q_{\mathrm{obs},i} - \overline{Q}_{\mathrm{obs}})^{2}}}
\label{eq:cor}
\end{equation}
\begin{equation}
\mathrm{FHV}\,(\%) = \frac{\sum_{h=1}^{H} (Q_{\mathrm{sim},h} - Q_{\mathrm{obs},h})}{\sum_{h=1}^{H} Q_{\mathrm{obs},h}} \times 100
\label{eq:fhv}
\end{equation}
\begin{equation}
\mathrm{FLV}\,(\%) = -1 \cdot \frac{\sum_{l=1}^{L} \left[ \log(Q_{\mathrm{sim},l}) - \log(Q_{\mathrm{sim},L}) \right] - \sum_{l=1}^{L} \left[ \log(Q_{\mathrm{obs},l}) - \log(Q_{\mathrm{obs},L}) \right]}{\sum_{l=1}^{L} \left[ \log(Q_{\mathrm{obs},l}) - \log(Q_{\mathrm{obs},L}) \right]} \times 100
\label{eq:flv}
\end{equation}
\begin{equation}
\mathrm{PBias}\,(\%) = \frac{\sum_{i=1}^{n} Q_{\mathrm{sim},i} - \sum_{i=1}^{n} Q_{\mathrm{obs},i}}{\sum_{i=1}^{n} Q_{\mathrm{obs},i}} \times 100
\label{eq:pbias}
\end{equation}

Second, we assess the hydrologic interpretability of PECAN's internal states and fluxes using external hydrologic references. Specifically, we first evaluate temporal consistency over the testing period (WY2013--WY2022) by computing spatially averaged daily and monthly fast-flow storage, slow-flow storage, and trash cell flux, and correlating them with PRISM precipitation, UA-SWE and SMAP surface and root zone soil moisture, and VIC-simulated ET. The daily analysis evaluates temporal correspondence at the model's native time step, whereas the monthly analysis highlights lower-frequency storage and seasonal relationships that may be obscured by event-scale variability. Because SMAP is only available from April 2015 onward, soil moisture diagnostics are restricted to the overlapping period within the testing window, whereas UA-SWE is available across the full testing period. We then plot the monthly climatology to examine whether the spatial distribution patterns of the internal states align with those of the external hydrologic reference datasets.

Third, we examine how the two mass conservation formulations (MC versus MCR) influence the representation of unresolved water sources and sinks. Specifically, we plot spatial monthly climatology of MR flux to characterize when and where relaxation is active (i.e., injecting additional water into the modeling system or producing additional water loss), and to understand how it operates alongside the trash cell mechanism. We also compare the trash cell flux in PECAN\textsubscript{MC} with its functional counterpart in PECAN\textsubscript{MCR}, defined as the combined contribution of trash cell and MR flux, to assess how relaxation changes the overall water loss simulation.

\section{Results}
\label{sec:results}

\subsection{Overall Model Performance}
\label{sec:overall_model_performance}

\subsubsection{Performance Metrics}
\label{sec:performance_metrics}

To provide an overview of simulation accuracy, Table~\ref{tab:performance} presents the performance of the lumped baseline models (LSTM, MC-LSTM, and MCR-LSTM) and the distributed PECAN variants (PECAN\textsubscript{MC} and PECAN\textsubscript{MCR}) during the testing period. The distributed PECAN models are evaluated at both the upstream gauge (G1: USGS 06614800) and the outlet gauge (G2: USGS 06620000), while the remaining baselines are evaluated at the outlet gauge only. For each metric and gauge, the best-performing value is highlighted in red, and the second-best value is highlighted in blue.

\begin{table}[htbp]
\centering
\caption{Overall performance statistics of lumped baseline models (LSTM, MC-LSTM, MCR-LSTM), VIC, and distributed PECAN (PECAN$_{\text{MC}}$ and PECAN$_{\text{MCR}}$) evaluated at two USGS gauges (G1: 06614800, upstream; G2: 06620000, outlet). Arrow $\uparrow$ denotes that values closer to 1 are optimal, whereas $\rightarrow 0$ denotes that values closer to 0 are optimal. For each metric and gauge, the best-performing value is shown in bold red, and the second-best value is shown in bold blue.}
\label{tab:performance}
\vspace{6pt}
\setlength{\tabcolsep}{3pt}
\resizebox{\textwidth}{!}{%
\begin{tabular}{lcccccccccccc}
\toprule
\multirow{2}{*}{\textbf{Model}} & \multicolumn{2}{c}{\textbf{NSE ($\uparrow$)}} & \multicolumn{2}{c}{\textbf{KGE ($\uparrow$)}} & \multicolumn{2}{c}{\textbf{COR ($\uparrow$)}} & \multicolumn{2}{c}{\textbf{PBias (\%) ($\rightarrow 0$)}} & \multicolumn{2}{c}{\textbf{FHV (\%) ($\rightarrow 0$)}} & \multicolumn{2}{c}{\textbf{FLV (\%) ($\rightarrow 0$)}} \\
\cmidrule(lr){2-3} \cmidrule(lr){4-5} \cmidrule(lr){6-7} \cmidrule(lr){8-9} \cmidrule(lr){10-11} \cmidrule(lr){12-13}
 & G1 & G2 & G1 & G2 & G1 & G2 & G1 & G2 & G1 & G2 & G1 & G2 \\
\midrule
\multicolumn{13}{l}{\textbf{Distributed PBM baseline}} \\
VIC & N/A & 0.78 & N/A & \textcolor{blue}{\textbf{0.88}} & N/A & 0.89 & N/A & 5.42 & N/A & $-$9.25 & N/A & 81.24 \\
\midrule
\multicolumn{13}{l}{\textbf{Lumped DL baselines (local)}} \\
LSTM & N/A & 0.73 & N/A & 0.80 & N/A & 0.86 & N/A & $-$8.33 & N/A & $-$10.89 & N/A & $-$2106.15 \\
MC-LSTM & N/A & 0.68 & N/A & 0.70 & N/A & 0.83 & N/A & $-$6.14 & N/A & $-$25.67 & N/A & $-$70.51 \\
MCR-LSTM & N/A & 0.57 & N/A & 0.71 & N/A & 0.76 & N/A & \textcolor{blue}{\textbf{$-$4.22}} & N/A & $-$20.01 & N/A & $-$184.70 \\
\midrule
\multicolumn{13}{l}{\textbf{Lumped DL baselines (regional)}} \\
LSTM & N/A & 0.76 & N/A & 0.59 & N/A & \textcolor{red}{\textbf{0.95}} & N/A & 30.06 & N/A & 29.55 & N/A & \textcolor{red}{\textbf{39.06}} \\
MC-LSTM & N/A & 0.80 & N/A & 0.78 & N/A & 0.92 & N/A & 16.64 & N/A & \textcolor{red}{\textbf{5.14}} & N/A & $-$144.8 \\
MCR-LSTM & N/A & \textcolor{red}{\textbf{0.85}} & N/A & \textcolor{blue}{\textbf{0.88}} & N/A & \textcolor{blue}{\textbf{0.93}} & N/A & 9.75 & N/A & \textcolor{blue}{\textbf{$-$6.74}} & N/A & $-$153.7 \\
\midrule
\multicolumn{13}{l}{\textbf{Distributed PECAN (ours)}} \\
PECAN$_{\text{MC}}$ & 0.77 & 0.76 & \textcolor{red}{\textbf{0.85}} & 0.81 & 0.88 & 0.87 & \textcolor{red}{\textbf{1.45}} & $-$5.93 & \textcolor{red}{\textbf{$-$0.39}} & $-$10.37 & 87.95 & $-$120.7 \\
PECAN$_{\text{MCR}}$ & \textcolor{red}{\textbf{0.86}} & \textcolor{blue}{\textbf{0.81}} & 0.83 & \textcolor{red}{\textbf{0.89}} & \textcolor{red}{\textbf{0.93}} & 0.90 & 4.07 & \textcolor{red}{\textbf{2.18}} & $-$11.92 & $-$10.69 & \textcolor{red}{\textbf{82.40}} & \textcolor{blue}{\textbf{$-$59.95}} \\
\bottomrule
\end{tabular}%
}
\end{table}

Overall, PECAN achieves better or competitive streamflow simulation performance relative to both process-based and lumped DL baselines. PECAN\textsubscript{MCR} outperforms the process-based VIC model across all metrics except for FHV. Compared with the locally trained lumped DL baselines, both PECAN variants show notable improvement over their lumped mass-conserving counterparts as well as the data-driven LSTM across NSE, KGE, COR, and PBias. 
The local LSTM exhibits an extreme FLV ($-2106.15$\%), which should be interpreted cautiously because the metric's logarithmic formulation is highly sensitive to near-zero flows \citep{frame2022deep}.
Regional training strengthens the mass-conserving lumped models and therefore provides a more demanding large-sample reference. Against this stronger reference, PECAN remains competitive, with performance differences depending on the metric. Specifically, at G2 (i.e., the outlet), the regional MC-LSTM and MCR-LSTM outperform PECAN\textsubscript{MC} and PECAN\textsubscript{MCR} in NSE, COR, and FHV. In contrast, PECAN\textsubscript{MC} and PECAN\textsubscript{MCR} outperform their corresponding lumped counterparts in KGE, PBias, and FLV. 
Notably, for the standard LSTM, regional training does not yield the same consistent benefit relative to local training, with KGE and the bias-related metrics (PBias and FHV) deteriorating. 
Together, these results show that PECAN retains strong outlet performance while extending the modeling capability from lumped discharge prediction to spatially distributed watershed representation.

Within PECAN, PECAN\textsubscript{MCR} generally outperforms PECAN\textsubscript{MC}, supporting the effectiveness of the mass relaxation mechanism. Specifically, PECAN\textsubscript{MCR} achieves better NSE, COR, and FLV at both gauges, as well as better KGE at the outlet gauge. In contrast, PECAN\textsubscript{MC} performs better in PBias at G1 and FHV at both gauges.

\subsubsection{Hydrograph Evaluation}
\label{sec:hydrograph_evaluation}

\begin{figure}[!t]
\centering
\includegraphics[width=\textwidth]{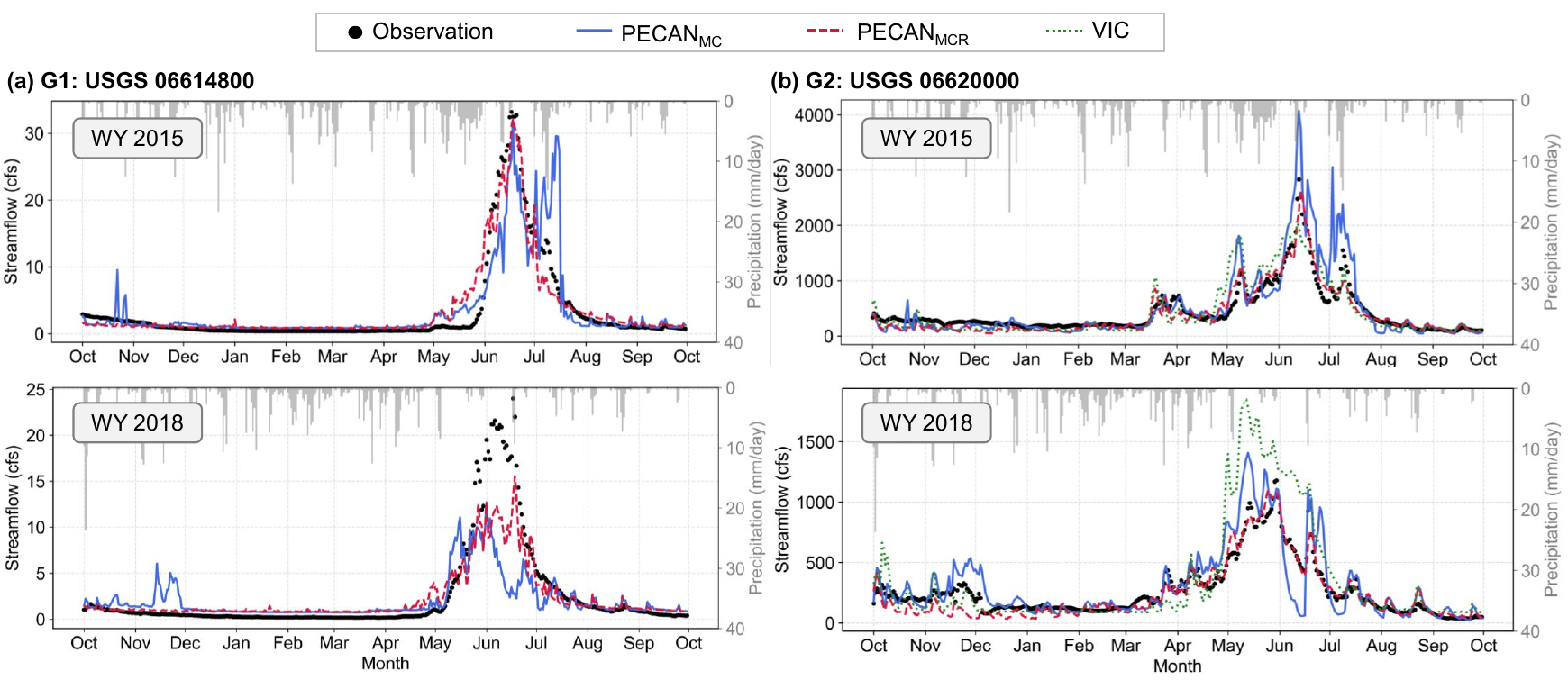}
\vspace{-0.7cm}
\caption{Representative hydrographs comparing observed and PECAN simulated streamflow at (a) the upstream station (G1: USGS 06614800) and (b) the outlet station (USGS 06620000). Results are shown for selected water years (WY 2015 and WY 2018). Observations are shown in black dots, PECAN\textsubscript{MC} in blue, and PECAN\textsubscript{MCR} in red dashed lines. VIC simulations are shown as green dots at the outlet station (G2) only. Gray bars represent daily precipitation on the right axis.}
\label{fig:hydrographs}
\end{figure}

To further assess how PECAN\textsubscript{MC} and PECAN\textsubscript{MCR} simulate streamflow dynamics, Figure~\ref{fig:hydrographs} shows representative hydrographs for selected water years (WY 2015 and WY 2018) at both the upstream (G1) and outlet (G2) gauges. VIC-simulated streamflow is also shown at G2 for comparison.

Both PECAN variants generally reproduce the snowmelt-driven hydrograph, with PECAN\textsubscript{MCR} better capturing the timing and magnitude of the runoff response. At G1, in WY 2015, PECAN\textsubscript{MCR} more closely follows the rising limb, peak, and subsequent recession, whereas PECAN\textsubscript{MC} incorrectly produces a delayed secondary peak in July. In WY 2018, both models underestimate the mid-June peak, but PECAN\textsubscript{MCR} shows a smaller underestimation and better reproduces the hydrograph shape during both the pre-June rising limb and the recession beginning in July.

At G2, where discharge magnitude is greatly larger than G1, the advantage of PECAN\textsubscript{MCR} remains evident. In WY 2015, PECAN\textsubscript{MCR} captures both the smaller May peak and the major June peak, whereas PECAN\textsubscript{MC} significantly overestimates both of them and VIC significantly underestimates the June peak. For the secondary peak in July, PECAN\textsubscript{MC} continues to overestimate, whereas both PECAN\textsubscript{MCR} and VIC underestimate it. In WY 2018, PECAN\textsubscript{MCR} closely follows the rising limb and recession, while PECAN\textsubscript{MC} exhibits larger fluctuations, alternating between over- and underestimation. VIC shows substantial overestimation during May and June around the seasonal peak.

\subsection{Internal State Comparison and Analysis}
\label{sec:internal_state_comparison}

\subsubsection{Temporal Correlation Analysis}
\label{sec:temporal_correlation_analysis}

\begin{figure}[!t]
\centering
\includegraphics[width=\textwidth]{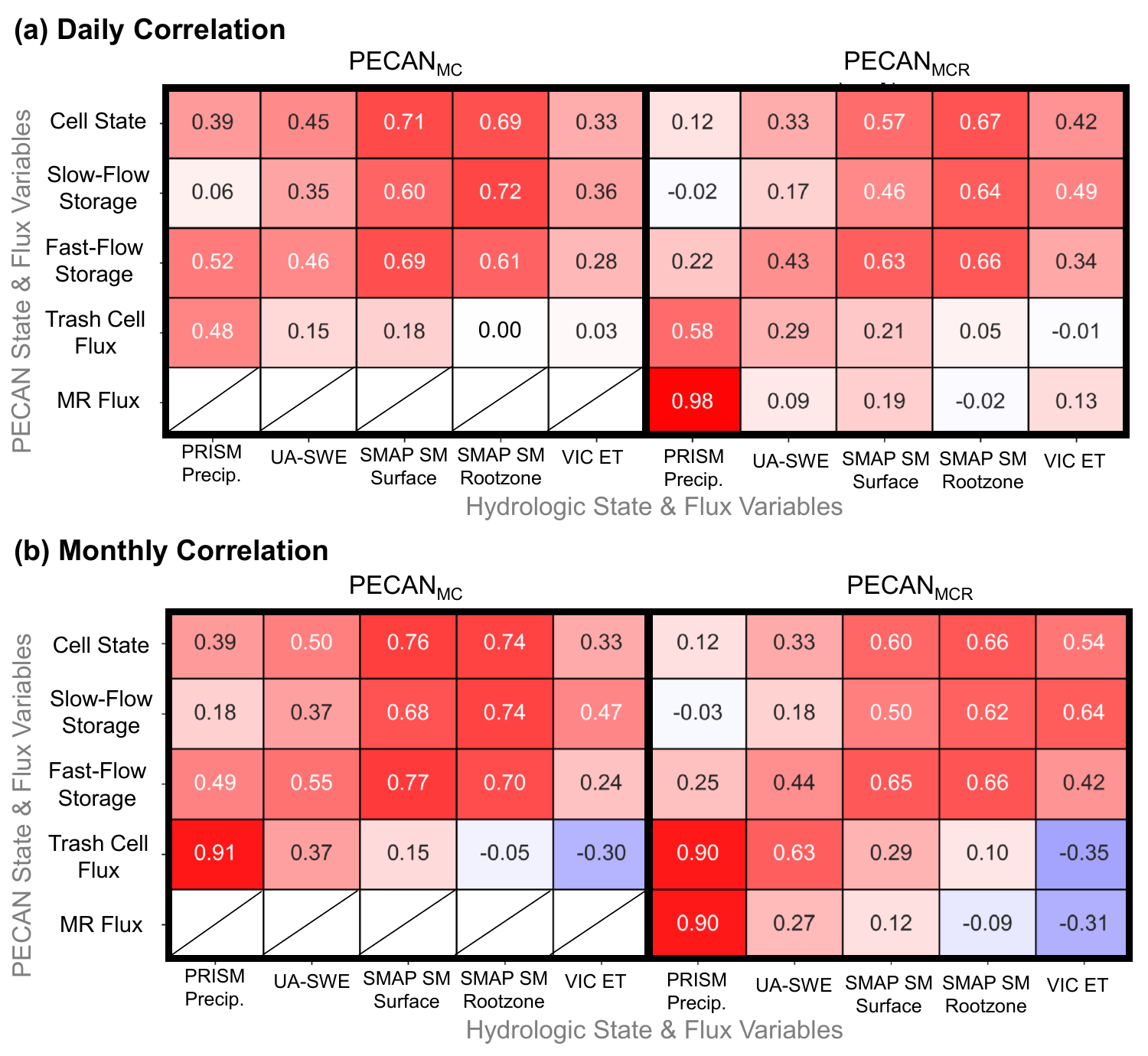}
\vspace{-0.7cm}
\caption{Temporal correlations between spatially averaged PECAN internal state/flux variables and hydrologic reference state/flux variables at the (a) daily and (b) monthly scales. PECAN internal variables include total cell state (combined fast-flow and slow-flow storage), individual storage components, trash cell flux, and MR flux (for PECAN\textsubscript{MCR} only). The hydrologic variables include PRISM precipitation, UA-SWE, SMAP soil moisture, and VIC ET. Correlation coefficients are shown within each grid cell, and color shading indicates correlation magnitude and sign (red for positive correlation and blue for negative correlation). Hatched cells indicate unavailable variable pairs.}
\label{fig:temporal_corr}
\end{figure}

A critical question in the design of PECAN is whether its internally learned states and fluxes, while being trained only on streamflow observations, encode physically meaningful hydrologic information. To address this question, Figure~\ref{fig:temporal_corr} shows the daily (panel (a)) and monthly (panel (b)) correlation of spatially averaged PECAN\textsubscript{MC} and PECAN\textsubscript{MCR} internal variables with PRISM precipitation, UA-SWE, SMAP soil moisture, and VIC ET.

Across both daily and monthly temporal scales, the total cell state (combined fast-flow and slow-flow storage) and its individual components exhibit strong positive correlations with SMAP soil moisture for both PECAN variants up to 0.8. By comparison, correlations with UA-SWE are lower, indicating weaker correspondence between the learned storage states and snowpack dynamics. For PECAN\textsubscript{MC}, the two storage components show a clear difference in correspondence with SMAP soil moisture: fast-flow storage correlates more strongly with SMAP surface soil moisture that responds more rapidly to dynamic meteorological forcings, whereas slow-flow storage correlates more strongly with SMAP root zone soil moisture. This difference is less clear in PECAN\textsubscript{MCR}, where both fast-flow and slow-flow storages show stronger correspondence with root zone soil moisture.

For the PECAN flux variables, the trash cell flux in both PECAN variants shows its strongest daily correlation with PRISM precipitation and weak or negligible correlations with SWE, SMAP soil moisture, and VIC ET. Similarly, the MR flux in PECAN\textsubscript{MCR} shows notably close daily synchronization with PRISM precipitation, with a correlation of 0.98. At the monthly scale, the trash cell flux becomes more strongly correlated with PRISM precipitation, with correlations exceeding 0.9 for both PECAN variants. Moreover, the correlation between trash cell flux and UA-SWE also increases at the monthly scale, reaching 0.37 for PECAN\textsubscript{MC} and 0.63 for PECAN\textsubscript{MCR}. By contrast, its relationships with SMAP soil moisture and VIC ET remain weak or become negative after monthly aggregation.

\subsubsection{Spatial Consistency Analysis}
\label{sec:spatial_consistency_analysis}

\begin{figure}[!htbp]
\centering
\includegraphics[width=\textwidth]{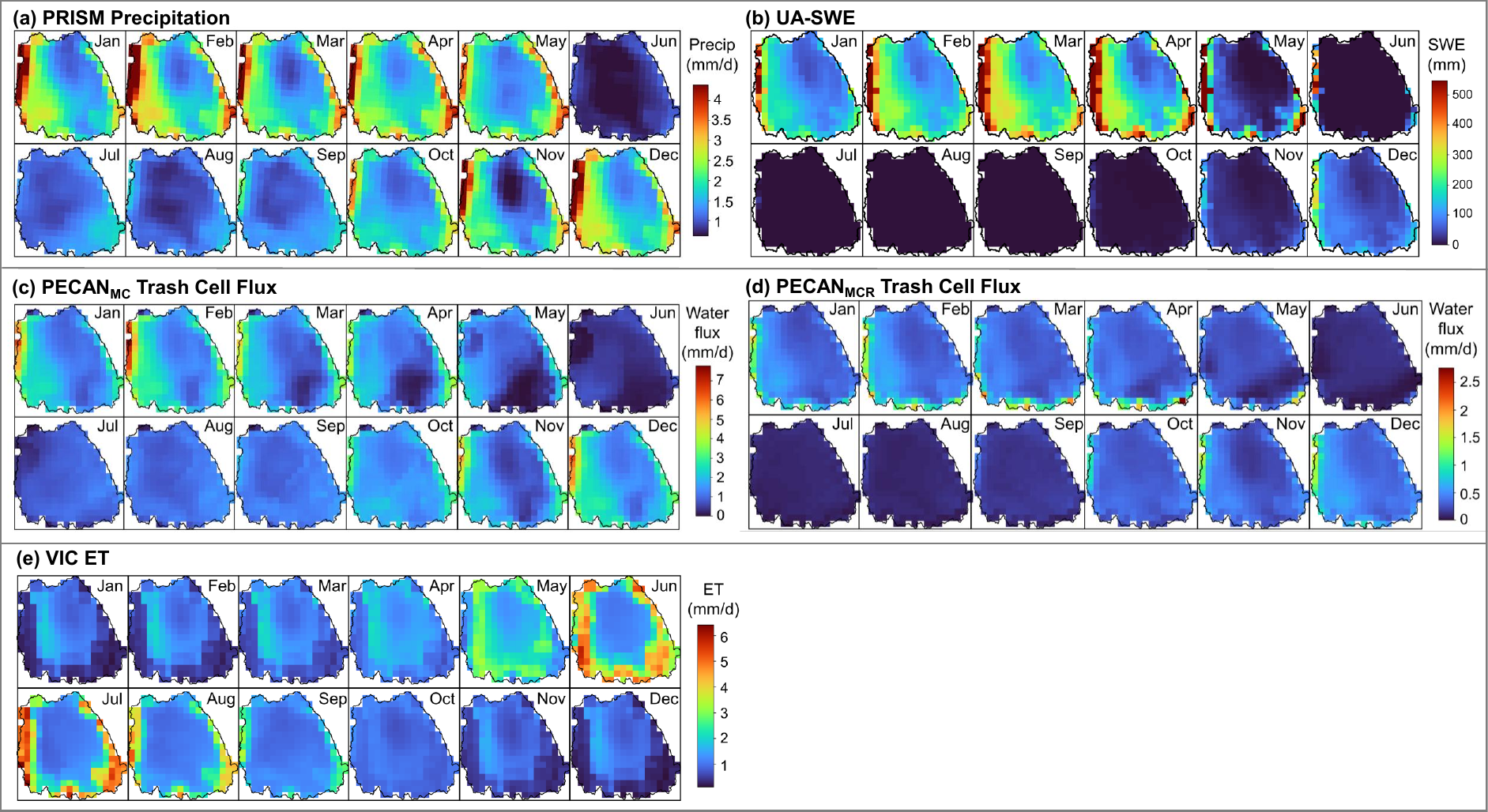}
\vspace{-0.7cm}
\caption{Spatial monthly climatology of (a) PRISM precipitation, (b) UA-SWE, (c) PECAN\textsubscript{MC} trash cell flux, (d) PECAN\textsubscript{MCR} trash cell flux, and (e) VIC-simulated ET.}
\label{fig:spatial_trash_cell}
\end{figure}

Building upon the temporal correlation results in Figure~\ref{fig:temporal_corr}, we further examine whether PECAN's learned states and fluxes show spatial patterns consistent with observation- and VIC-simulation-based hydrologic references. Figure~\ref{fig:spatial_trash_cell} examines the spatial consistency of PECAN trash cell flux with PRISM precipitation, VIC-simulated ET, and UA-SWE. These variables provide complementary hydrologic references for interpreting the trash cell flux: precipitation represents direct water input that exhibits strong temporal correspondence with trash cell shown in Figure~\ref{fig:temporal_corr}; VIC ET represents part of the water loss that the trash cell is intended to approximate; and UA-SWE represents accumulated snow storage, whose correspondence with trash cell flux becomes stronger at the seasonal scale.

PECAN trash cell fluxes spatially follow the precipitation and SWE patterns. PRISM precipitation and UA-SWE both show high-value regions along the western and eastern high-elevation ridges, indicating that snow storage is strongly organized by the spatial distribution of cold-season precipitation. PECAN\textsubscript{MC} trash cell flux generally mirrors this spatial structure. Higher trash cell flux clusters most prominently along the western and eastern boundaries, whereas lower values occur in the basin interior. PECAN\textsubscript{MCR} shows a similar but weaker pattern, with substantially smaller trash cell magnitudes than PECAN\textsubscript{MC}. This reduction occurs because PECAN\textsubscript{MCR} includes the additional MR flux, which provides another pathway for representing water removal. This MR mechanism is analyzed in detail in Section~\ref{sec:mass_conservation_relaxation}.

The spatial maps also reveal a seasonal offset between trash cell flux and SWE, providing additional context for their temporal relationship shown in Figure~\ref{fig:temporal_corr}. In both PECAN\textsubscript{MC} and PECAN\textsubscript{MCR}, trash cell flux begins to increase in November, peaks around February, and then declines. However, SWE continues to accumulate through winter and reaches its maximum around April before melting. VIC-simulated ET provides a further spatial and temporal contrast to the trash cell flux. During winter and spring, ET is lowest along the western and eastern high-elevation ridges where snow accumulates, while relatively higher ET occurs in the basin interior. During summer, ET increases along the western and eastern regions, whereas PECAN trash cell fluxes are nearly absent.

\begin{figure}[!t]
\centering
\includegraphics[width=\textwidth]{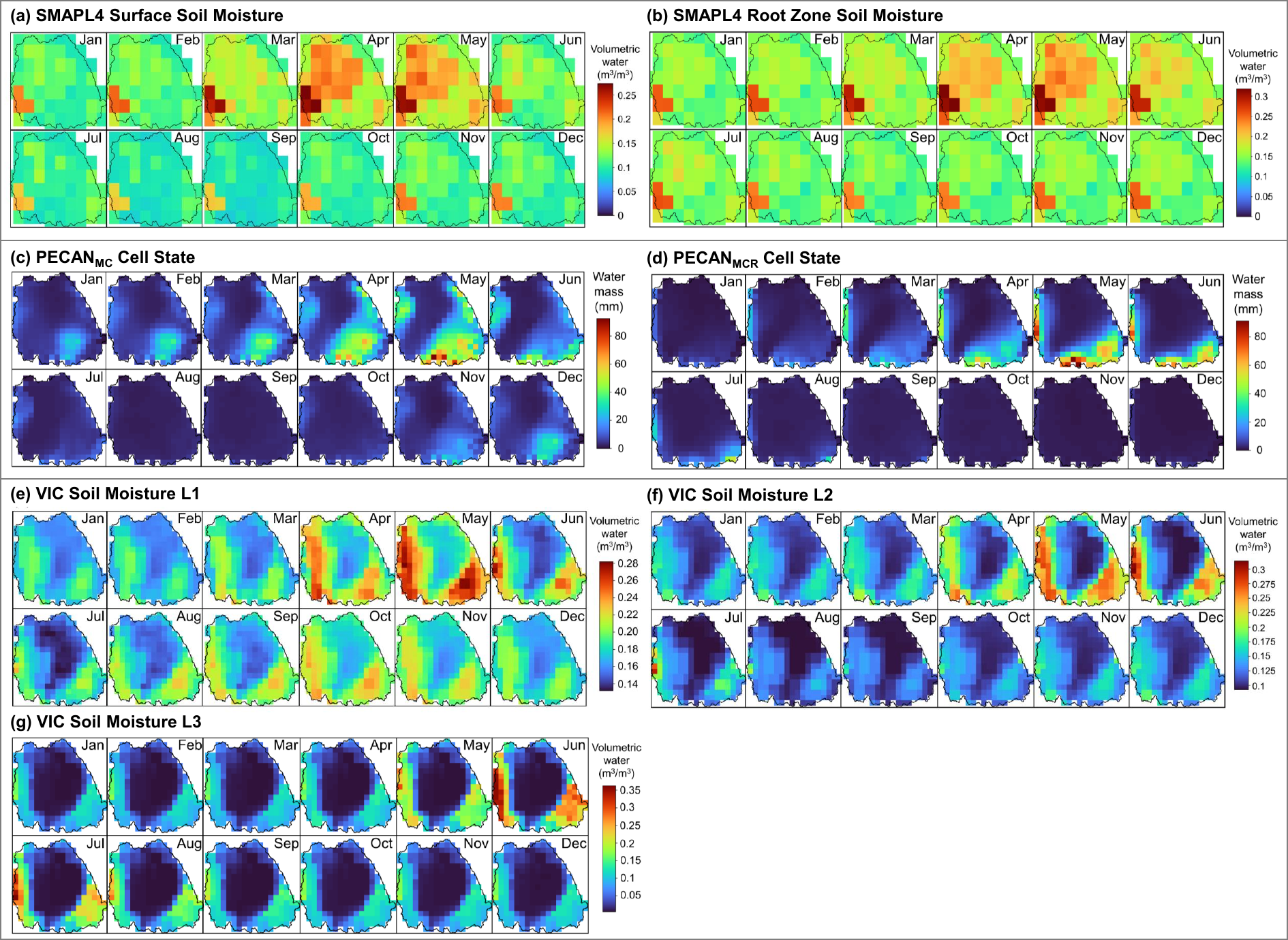}
\vspace{-0.7cm}
\caption{Spatial monthly climatology of (a) SMAP L4 surface soil moisture, (b) SMAP L4 root zone soil moisture, (c) PECAN\textsubscript{MC} cell state, (d) PECAN\textsubscript{MCR} cell state, and (e--g) VIC-simulated soil moisture layers 1--3. PECAN cell state is computed as the sum of fast-flow and slow-flow storage.}
\label{fig:spatial_cell_state}
\end{figure}

Figure~\ref{fig:spatial_cell_state} further compares monthly climatology of PECAN cell states, SMAP surface and root zone soil moisture, as well as three VIC-simulated soil moisture layers as an additional set of references. For clarity, only the total PECAN cell state (i.e., the sum of fast-flow and slow-flow storage) is shown here; the individual storage components are provided in Figures S1 and S2 in the Appendix.

Overall, the spatial distributions of PECAN cell states differ substantially from both SMAP- and VIC-simulated soil moisture, although the differences are relatively smaller compared with VIC. In SMAP, both surface and root zone soil moisture exhibit a persistent moisture maximum in the southwestern portion of the watershed. During the melt season from March to May, elevated soil moisture expands from this southwestern region toward the northwestern interior region, while a second wet area also develops in the southeastern portion of the watershed. In contrast, both PECAN\textsubscript{MC} and PECAN\textsubscript{MCR} show their main storage hotspots along the western ridge and in the southeastern portion of the watershed. These hotspot locations are more consistent with VIC-simulated soil moisture across the three soil layers.

Nevertheless, the two PECAN variants also differ in the seasonal and spatial concentration of their storage hotspots. In PECAN\textsubscript{MC}, the southeastern hotspot appears in winter and strengthens during spring, while storage along the western ridge also becomes more pronounced. In PECAN\textsubscript{MCR}, the southeastern signal is weak in winter, and the storage hotspots remain more spatially concentrated during the melt season: the southeastern hotspot does not extend as broadly toward the eastern ridge, and the western-ridge hotspot appears as a narrower strip.

\subsection{Mass Conservation and Relaxation in PECAN}
\label{sec:mass_conservation_relaxation}

\begin{figure}[!t]
\centering
\includegraphics[width=\textwidth]{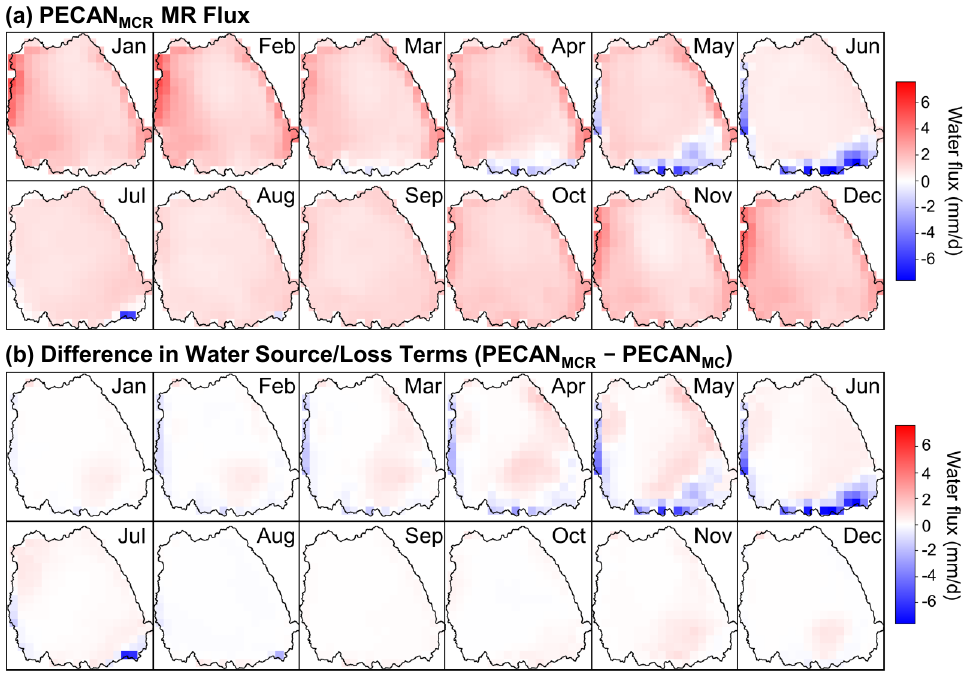}
\vspace{-0.7cm}
\caption{Spatial monthly climatology of (a) PECAN\textsubscript{MCR} MR flux, and (b) the difference in source/loss terms between PECAN\textsubscript{MCR} and PECAN\textsubscript{MC}, defined as PECAN\textsubscript{MCR} trash cell flux + PECAN\textsubscript{MCR} MR flux -- PECAN\textsubscript{MC} trash cell flux. In panel (a), positive values (red) indicate water removal through the MR pathway, whereas negative values (blue) indicate water injection. In panel (b), positive values (red) represent greater water removal in PECAN\textsubscript{MCR} relative to PECAN\textsubscript{MC}, and negative values (blue) represent reduced removal or net water injection.}
\label{fig:mr_flux}
\end{figure}

The MR gate in PECAN\textsubscript{MCR} modifies the strict precipitation-based MC formulation in PECAN\textsubscript{MC}, thereby changing how unresolved water sources and sinks are represented within the modeling system. To understand the functionality of the MR mechanism, we examine the MR flux itself and compare how the combined source/loss representation differs between the two PECAN variants, as shown in Figure~\ref{fig:mr_flux}.

Figure~\ref{fig:mr_flux}(a) shows the monthly climatology of MR flux in PECAN\textsubscript{MCR}. MR flux exhibits a seasonal cycle broadly similar to that of the precipitation and trash cell flux shown in Figure~\ref{fig:spatial_trash_cell}. Specifically, beginning in the fall (i.e., October), water loss initiates along the western and eastern ridges. This loss intensifies through the winter months (November to February) before weakening in early spring (March).~In contrast to the trash cell flux, however, the MR flux subsequently becomes negative, indicating water injection. Negative MR flux first emerges in the southern portion of the basin, extends toward the western ridge in May, intensifies across both regions in June, and largely diminishes by July. Outside the April--July period, the MR pathway primarily functions as an additional water-removal mechanism that complements the trash cell.

Figure~\ref{fig:mr_flux}(b) shows the monthly climatology difference between the water source/loss terms in PECAN\textsubscript{MCR} (trash cell flux + MR flux) and PECAN\textsubscript{MC} (trash cell flux only). Differences are relatively small during late summer through winter, but become pronounced from February through July, with the largest contrasts occurring during April--June. This pattern suggests that MC and MCR behaviors diverge mostly during snowmelt. Specifically, beginning in February, PECAN\textsubscript{MCR} shows reduced water removal relative to PECAN\textsubscript{MC} (blue areas) along the western ridge, while greater removal emerges across the southeastern portion of the watershed. Such spatial contrast strengthens through spring. By May and June, reduced removal or net water injection becomes particularly pronounced along the western and southern watershed boundary, whereas more water removal occurs primarily across the central interior and northeastern ridge. The magnitude of these differences decreases substantially by July and becomes minimal from August onward. Overall, the two mass conservation formulations have similar water loss behavior for most of the year, with MR primarily retaining or adding water during the spring melt period rather than systematically increasing or decreasing overall water removal.

\section{Discussion}
\label{sec:discussion}

\subsection{Physical Rationale of Mass-Conserving Structures in Hydrologic Deep Learning}
\label{sec:physical_rationale_mass}

Mass-conserving recurrent architectures such as MC-LSTM and MCR-LSTM are motivated by a storage--flux representation of hydrologic systems. Specifically, the LSTM cell states are assigned to the role of water storage, the gating mechanisms regulate water partitioning and release, and state transitions are constrained to follow a discrete mass balance equation. Through this structural reformulation, the abstract, high-dimensional recurrent dynamics of an LSTM resemble the familiar, intuitive bucket-like framework analogous to conceptual hydrologic models, representing a meaningful step toward improving the transparency of DL models in hydrology.

However, physical validity is not an inherent property of the analogy itself, but also depends on~how accurately the analogy is implemented~to represent real-world physics (i.e., hydrological processes in this case). When essential mass balance components (e.g., unobserved water inputs beyond precipitation) are ignored, the prescribed flux partition becomes structurally incomplete. Consequently, the mass-conserving architecture is forced to search within a solution space that deviates from physical reality, leading to deteriorated simulation accuracy relative to unconstrained LSTM \citep{frame2023strictly,wang2025investigating,wang2025mcr,wi2024need}. The performance degradation highlights a research need to understand how analogy-prescribed physics interacts with data-driven learning at a mechanistic level, and to inform the design of future physics-aware DL architectures. At present, this interaction remains insufficiently clarified, limiting progress toward reconciling deep learning with hydrologic reasoning.

The development of PECAN is therefore motivated by this knowledge gap. PECAN extends the storage--flux formulation from lumped to spatially distributed hydrologic modeling and provides two alternative conservation implementations: a precipitation-based mass conservation (MC) and a mass conservation relaxed (MCR) variant that introduces a bidirectional pathway for additional water sources beyond precipitation. The results in Section~\ref{sec:overall_model_performance} first establish that PECAN is an effective distributed hydrologic model: both variants outperform the locally trained lumped counterparts and remain competitive with substantially stronger regionally trained lumped baselines that additionally use static catchment attributes that are not provided to PECAN. In addition, PECAN\textsubscript{MCR} outperforms the process-based VIC model. More importantly, PECAN is intended not only for accurate streamflow simulation, but also for hydrologic diagnosis. Specifically, its contrasting MC and MCR formulations can be used as a framework for examining how grid-level latent states, fluxes, and compensatory mechanisms emerge under spatial heterogeneity. In the following Sections~\ref{sec:hydrologic_interpretation_pecan}--\ref{sec:role_mass_relaxation}, we leverage this latent information to analyze how PECAN allocates mass across space and time, and to revisit the mechanistic tensions between physical constraints and data-driven learning.

\subsection{Hydrologic Interpretation of PECAN Cell States with Soil Moisture}
\label{sec:hydrologic_interpretation_pecan}

Because the PECAN cell state is formulated as a mass-conserving storage variable, soil moisture provides a natural reference for evaluating whether the learned states capture meaningful terrestrial water-storage dynamics. Results in Section~\ref{sec:internal_state_comparison} show encouraging temporal agreement with SMAP soil moisture. Both PECAN variants exhibit strong correlations with SMAP surface and root-zone soil moisture at daily and monthly scales (Figure~\ref{fig:temporal_corr}), indicating that the learned cell states capture both short-term variability and the broader seasonal evolution of watershed wetness.

However, the spatial agreement is considerably weaker. PECAN cell states differ substantially from SMAP soil moisture in the location and extent of high-storage regions (Figure~\ref{fig:spatial_cell_state}). Several characteristics of the SMAP L4 product warrant caution when interpreting this disagreement. First, SMAP L4 is provided at approximately 9-km resolution, which can smooth spatial variability within this relatively small and topographically heterogeneous watershed compared with the 4-km PECAN grid. In addition, SMAP brightness-temperature observations have little or no immediate impact on L4 soil-moisture estimates under frozen or snow-covered conditions, causing the L4 estimates to rely more heavily on the underlying land-surface model and data-assimilation system \citep{reichle2019version}. Consequently, the resulting SMAP spatial patterns can also depend on the model formulation and meteorological forcing used in the assimilation system, making SMAP an informative but not definitive reference for evaluating PECAN cell states.

VIC-simulated soil moisture provides supporting evidence that the weak spatial agreement between PECAN and SMAP may partly reflect uncertainty in the SMAP spatial representation. Both our calibrated PRISM-driven VIC simulation and the independently developed NLDAS-driven VIC product of \citet{xia2012continental} differ substantially from SMAP, while showing broad spatial patterns relatively more similar to PECAN, including elevated storage along portions of the western ridge and southeastern watershed (Figure~\ref{fig:spatial_cell_state}; Figure S3 in the Appendix). Taken together, PECAN shows promising correspondence with soil moisture dynamics over time, while stronger post hoc interpretation across space requires further evaluation across additional watersheds.

\subsection{Representation of Snow Storage and Release in PECAN}
\label{sec:representation_snow_storage}

SWE also represents a major terrestrial water-storage component. However, compared with soil moisture, PECAN cell states show much weaker correspondence with SWE both temporally and spatially (Figures~\ref{fig:temporal_corr}--\ref{fig:spatial_cell_state}). Instead, SWE shares seasonal variability partially with both the PECAN cell state and trash cell flux (Figure S4 in the Appendix), indicating that the snow-related signal is not represented by the learned storage state alone but is also expressed through the water loss pathway. Unlike soil moisture, which is more dynamically coupled with runoff generation over short timescales, snow can remain stored for months before being released during melt. This difference suggests that PECAN cell states tend to capture storage with a relatively synchronous runoff response, while the delayed storage--release behavior is not explicitly resolved.

The trash cell behavior provides further insight into how this unresolved snow-related signal is represented. Although the trash cell flux is intended to represent water losses (e.g., ET, snow sublimation, deep percolation, or lateral export out of the modeled domain), it shows weak or reversed correlation with VIC-simulated ET and instead corresponds with precipitation---and, consequently, SWE accumulation---more strongly, especially at the monthly scale (Figures~\ref{fig:temporal_corr} and~\ref{fig:spatial_trash_cell}). This suggests that part of the incoming water associated with the seasonal snow cycle is expressed through the trash cell pathway rather than through a persistent snow-like storage state. As a result, the trash cell ends up not only representing genuine water losses but also part of the unresolved snow-related water that is not retained within the learned storage states. This unresolved seasonal storage--release behavior motivates examining whether mass relaxation provides an additional compensatory mechanism, as discussed in the following Section~\ref{sec:role_mass_relaxation}.

\subsection{Role of Mass Relaxation in Snow-Dominated Regimes and Its Implications}
\label{sec:role_mass_relaxation}

Results in Section~\ref{sec:overall_model_performance} show that PECAN\textsubscript{MCR} generally outperforms PECAN\textsubscript{MC} across evaluation metrics (Table~\ref{tab:performance}), and more accurately reproduces the rising limb, peak flow, and recession of the hydrograph at both gauges (Figure~\ref{fig:hydrographs}). Similar behavior has been reported for lumped mass-conserving architectures in \citet{wang2025mcr}, where the MCR-LSTM improves both MC-LSTM and LSTM in the snow-dominated South Rockies where our study watershed is located. Our distributed results are consistent with this previous finding. At the same time, PECAN's spatially resolved states and fluxes provide additional information for examining how and where mass relaxation operates, motivating a closer examination of the mechanisms underlying its benefit in snow-dominated regimes.

Given the mass relaxation behavior presented in Section~\ref{sec:mass_conservation_relaxation}, the improved simulation of PECAN\textsubscript{MCR} is likely linked to its ability to ``inject'' water during the snowmelt season. Specifically, both PECAN variants exhibit the broader difficulty in representing delayed snow storage and release identified in Section~\ref{sec:representation_snow_storage}. The key difference is that, when the learned cell state contains insufficient water during snowmelt, PECAN\textsubscript{MC} has no additional pathway to increase water availability. In contrast, the mass relaxation pathway in PECAN\textsubscript{MCR} can reintroduce water during May and June (Figure~\ref{fig:mr_flux}), providing flexibility to partially offset this seasonal storage deficit. In this sense, mass relaxation extends its original role of accommodating unobserved water inputs beyond precipitation \citep{wang2025mcr}, while also acting here as a learned compensation for unrepresented hydrologic processes. We note that a similar snow-related compensation mechanism has also been reported in a large-sample lumped mass-conserving perceptron study in \citet{wang2025towards}.

\begin{figure}[!htbp]
\centering
\includegraphics[width=\textwidth]{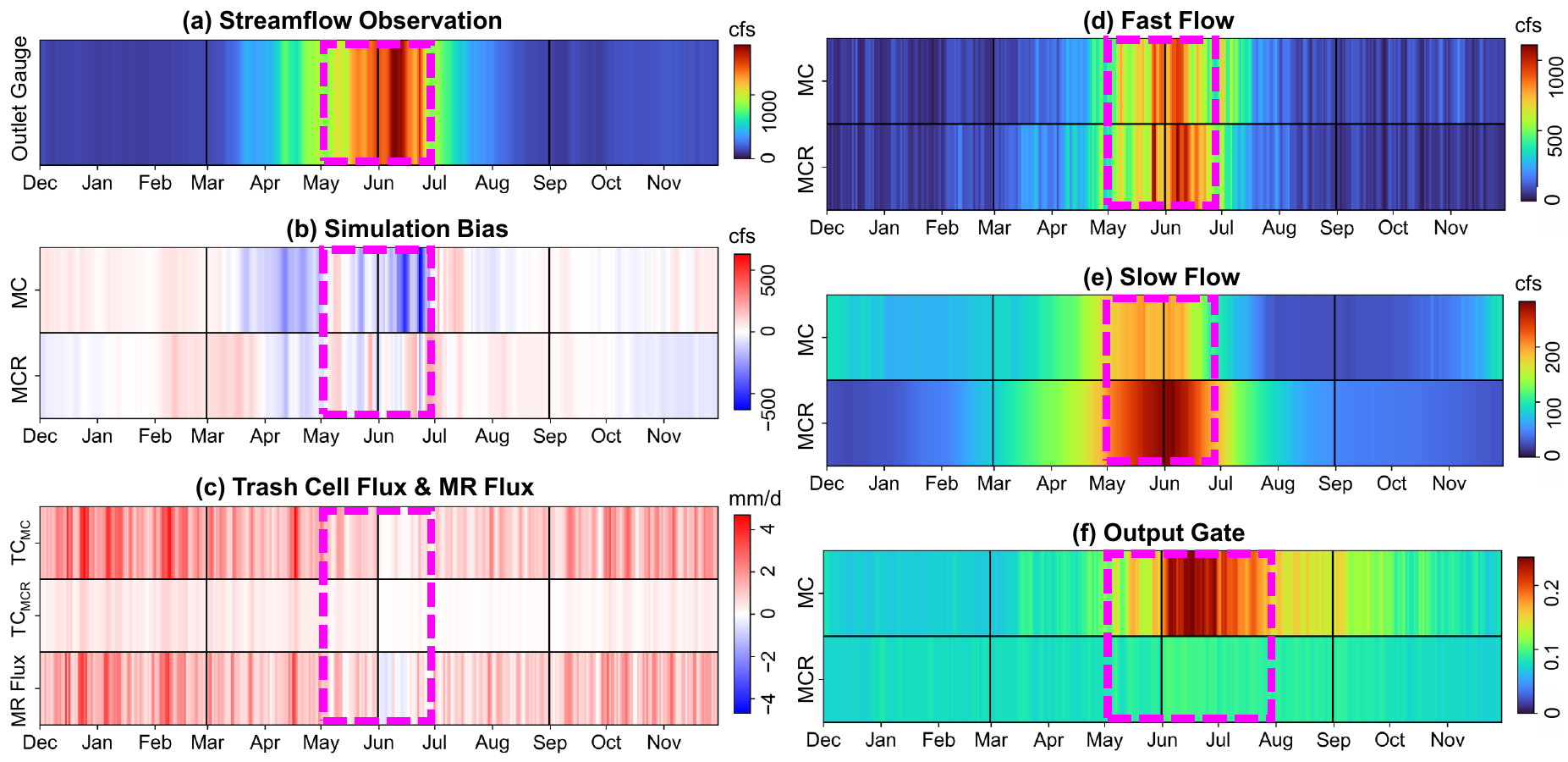}
\vspace{-0.7cm}
\caption{Climatological daily mean of (a) observed streamflow, (b) PECAN simulation bias, and (c-f) PECAN internal variables. For each calendar day of the year, values are averaged across all years within the testing period, and the resulting daily climatology is plotted from December to November (DJF-first). Panels compare PECAN\textsubscript{MC} (labeled as MC) and PECAN\textsubscript{MCR} (labeled as MCR): (c) trash cell flux for MC and MCR (TC\textsubscript{MC} and TC\textsubscript{MCR}) and MR Flux for MCR only, (d) fast flow, (e) slow flow, and (f) output gate. Vertical black solid lines mark the boundaries between DJF (winter), MAM (spring), JJA (summer), and SON (fall). Pink dashed boxes highlight the snowmelt-driven high-flow period (May--June/July) for visual comparison across panels.}
\label{fig:melt_climatology}
\end{figure}

To further test this interpretation, we examine the climatological daily streamflow, simulation bias, internal fluxes, and output gate behavior of the two PECAN variants during the snowmelt season, as shown in Figure~\ref{fig:melt_climatology}. According to the outlet streamflow and simulation bias in Figures~\ref{fig:melt_climatology}a and~\ref{fig:melt_climatology}b, PECAN\textsubscript{MC} substantially underestimates streamflow during the melt period (highlighted by the pink boxes), particularly in June, whereas PECAN\textsubscript{MCR} markedly reduces the bias. The improved bias coincides with water injection through the MR flux during May and June (blue strips in the last row of Figure~\ref{fig:melt_climatology}c). Importantly, PECAN\textsubscript{MC} already reduces its trash cell flux to nearly zero (TC\textsubscript{MC}, top row of Figure~\ref{fig:melt_climatology}c) during this period and increases its output gate substantially (Figure~\ref{fig:melt_climatology}f), indicating that the model is neither continuing to discard substantial water nor strongly restricting its release. Yet PECAN\textsubscript{MC} still produces less runoff through both the fast flow and slow flow pathways as compared to PECAN\textsubscript{MCR} (Figures~\ref{fig:melt_climatology}d and~\ref{fig:melt_climatology}e). Taken together, the melt season underestimation of PECAN\textsubscript{MC} is primarily caused by limited water availability rather than by restricted release, supporting our interpretation that the MR flux improves snowmelt simulation by alleviating the storage water deficit.

\subsection{Limitations, Implications, and Future Directions}
\label{sec:limitations_implications_future}

The difficulty in representing delayed snow storage and release in PECAN may partly reflect a structural feature of the sequence-to-one training framework commonly used in hydrologic DL models. While the water-bucket concept discussed in Section~\ref{sec:physical_rationale_mass} provides a physical motivation and design target, it is not fully realized under the current sequence-to-one training framework. In a conventional bucket model, storage evolves continuously through time, with the state from one time step carried forward to the next. In contrast, during DL training, each input window is treated as an independent sample, and internal states are reset at the start of the sequence. Consequently, the DL model is not asked to maintain a single, continuous storage trajectory over the full simulation period; it only needs to close the mass balance (if constraints are imposed) and reproduce streamflow within each window.

Consequently, DL model behavior depends on how much antecedent information can be reconstructed from a finite history, consistent with prior findings that LSTM-based performance is sensitive to the chosen sequence length \citep{gauch2021proper,hashemi2022regime}. When the relevant hydrologic memory is not sufficiently captured within this finite history, longer-term processes may be underrepresented, which may contribute to the difficulty of both PECAN variants in representing delayed snow storage and release. This tension between sequence-based training and the continuous-state assumption underlying the bucket analogy has received increasing attention in recent literature \citep{acunaespinoza2024bucket,delafuente2023towards,poudel2026parsimony}. Accordingly, the learned states in DL models such as PECAN should primarily be understood as hydrologically informed latent representations rather than presumed to directly represent specific physical hydrologic variables without further validation.

A second limitation is that the present analysis focuses on a single snow-dominated watershed. Future work could apply the same analysis across watersheds with different hydroclimatic conditions and runoff-generation mechanisms to reveal how PECAN's internal dynamics vary across hydrologic regimes and which behaviors are generalizable. This broader assessment will help provide a more general scientific understanding of how physics-aware DL represents various watershed dynamics.

Moving forward, we believe that realizing a physically meaningful water bucket analogy in hydrologic DL models will require advances along several complementary avenues. First, reducing or eliminating state resets across input windows would promote continuous state evolution and resemble real catchment dynamics more closely, potentially allowing the learned states to move beyond task-driven effective storage representation. Second, future development should incorporate more complete process representations, such as explicit snow and energy-balance components as also highlighted by recent studies \citep{song2026physics,wang2025towards,xu2022hybrid}, rather than relying on continuous and long antecedent information alone for the model to autonomously learn these seasonal dynamics. This need is also indicated by the comparison between PECAN\textsubscript{MCR} and PECAN\textsubscript{MC}: despite the longer input window used by PECAN\textsubscript{MCR}, its learned cell states do not exhibit a correspondingly clearer or more persistent snow-like storage representation than those of PECAN\textsubscript{MC}. Third, further progress may also explore whether mass conservation can be integrated into other distributed recurrent sequence models (e.g., WaveCastNet in \citealp{lyu2025rapid}), and whether alternative gating formulations can provide a more appropriate representation of storage evolution and water flux partitioning.

\section{Conclusions}
\label{sec:conclusions}

In this study, we introduced PECAN, a distributed physics-aware deep learning framework that combines a mass-conserving ConvLSTM-based runoff module with a physically based routing module for hydrologic modeling. We implement two alternative conservation formulations in PECAN, namely a precipitation-based mass conservation constraint (PECAN\textsubscript{MC}), and a relaxed conservation constraint that allows additional input water sources beyond precipitation (PECAN\textsubscript{MCR}). Both PECAN variants are trained using only streamflow observations at one upstream and one outlet gauge in a snow-dominated headwater HUC-8 basin of the North Platte River in Colorado. We evaluate their performance at the basin outlet against lumped mass-conserving counterparts (MC-LSTM and MCR-LSTM), a standard LSTM, as well as the process-based VIC model. We further interpret PECAN's internal states and fluxes using independent hydrologic references, including soil moisture, SWE, and VIC-simulated ET, and investigate the hydrologic implications of the mass relaxation mechanism. The main conclusions are as follows:

\begin{enumerate}
    \item PECAN provides competitive streamflow simulation relative to lumped mass-conserving baselines, while simultaneously providing spatially explicit states and fluxes for hydrologic diagnosis. Notably, PECAN\textsubscript{MCR} achieves accuracy on par with the process-based VIC model, highlighting the effectiveness of imposing mass conservation constraints on a deep learning model in a spatially distributed runoff and routing framework.
    \item PECAN reveals a limitation in representing seasonal snow storage and delayed release under a formulation without an explicit snow accumulation and melt component. PECAN cell states show strong temporal correspondence with soil moisture but much weaker temporal and spatial agreement with SWE, suggesting that they more readily represent storage with a relatively synchronous runoff response than delayed snow storage and release. The trash cell flux, meanwhile, follows the spatiotemporal pattern of precipitation and seasonal SWE, suggesting that a portion of the snow signal is treated as unresolved water mass in PECAN and is discarded rather than retained for later snowmelt.
    \item Mass relaxation in PECAN\textsubscript{MCR} acts as a correction mechanism to compensate for the underrepresented seasonal snow storage and release. Specifically, MR reintroduces water into the cell state during the snowmelt period, supplying the additional water mass needed to reproduce snowmelt-driven streamflow and improving peak flow simulation compared to PECAN\textsubscript{MC}.
\end{enumerate}

To conclude, PECAN demonstrates that spatially distributed mass-conserving neural architectures can deliver strong simulation performance, while offering richer latent storages and fluxes that complement lumped models for understanding model behavior and assessing the realism of imposed constraints. At the same time, the results highlight the opportunities to advance physics-aware DL in hydrology, including preserving more continuous state evolution across training samples, incorporating more complete hydrologic process representations, and extending physically constrained design principles to other distributed neural architectures. As a proof of concept in a snow-dominated watershed, this study illustrates that PECAN's internal state diagnostics can expose specific structural limitations in a model. Applying PECAN across broader hydroclimatic regimes could help identify more missing processes (e.g., groundwater--surface water interactions in karst terrain), thereby informing the design of physical constraints and providing new insights into the synergy between physical theory and data-driven learning.

\section*{Acknowledgments}

The financial support for this work is made available from the National Science Foundation (NSF) CAREER Award (No. 2618864) and the U.S. Bureau of Reclamation (USBR) Project No. R24AC00032. We also thank the National Oceanic and Atmospheric Administration (NOAA)'s Climate Program Office, CVP and MAPP programs (Award Number: NA23OAR4310459) for partial financial support.

\section*{Code Availability}
The source code will be made publicly available upon publication.

\bibliography{references}
\bibliographystyle{iclr2025_conference}


\clearpage
\appendix
\section*{Appendix}

\setcounter{figure}{0}
\setcounter{table}{0}
\renewcommand{\thefigure}{S\arabic{figure}}
\renewcommand{\thetable}{S\arabic{table}}
\renewcommand{\theHfigure}{S\arabic{figure}}
\renewcommand{\theHtable}{S\arabic{table}}

\begin{figure}[!ht]
\centering
\includegraphics[width=\textwidth]{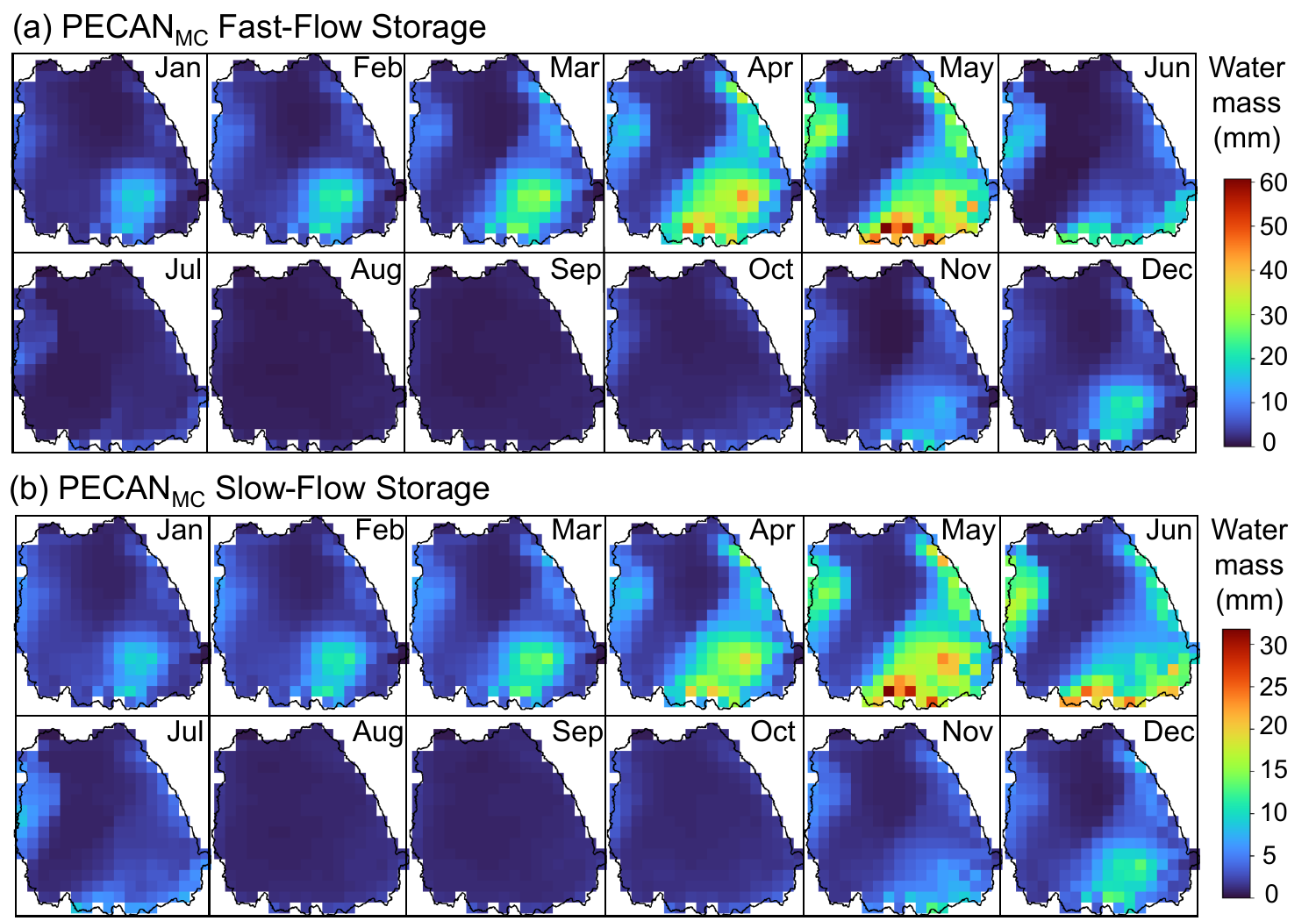}
\vspace{-0.7cm}
\caption{Spatial monthly climatology of (a) PECAN\textsubscript{MC} fast-flow storage and (b) PECAN\textsubscript{MC} slow-flow storage.}
\label{fig:S1}
\end{figure}

\clearpage
\begin{figure}[!ht]
\centering
\includegraphics[width=\textwidth]{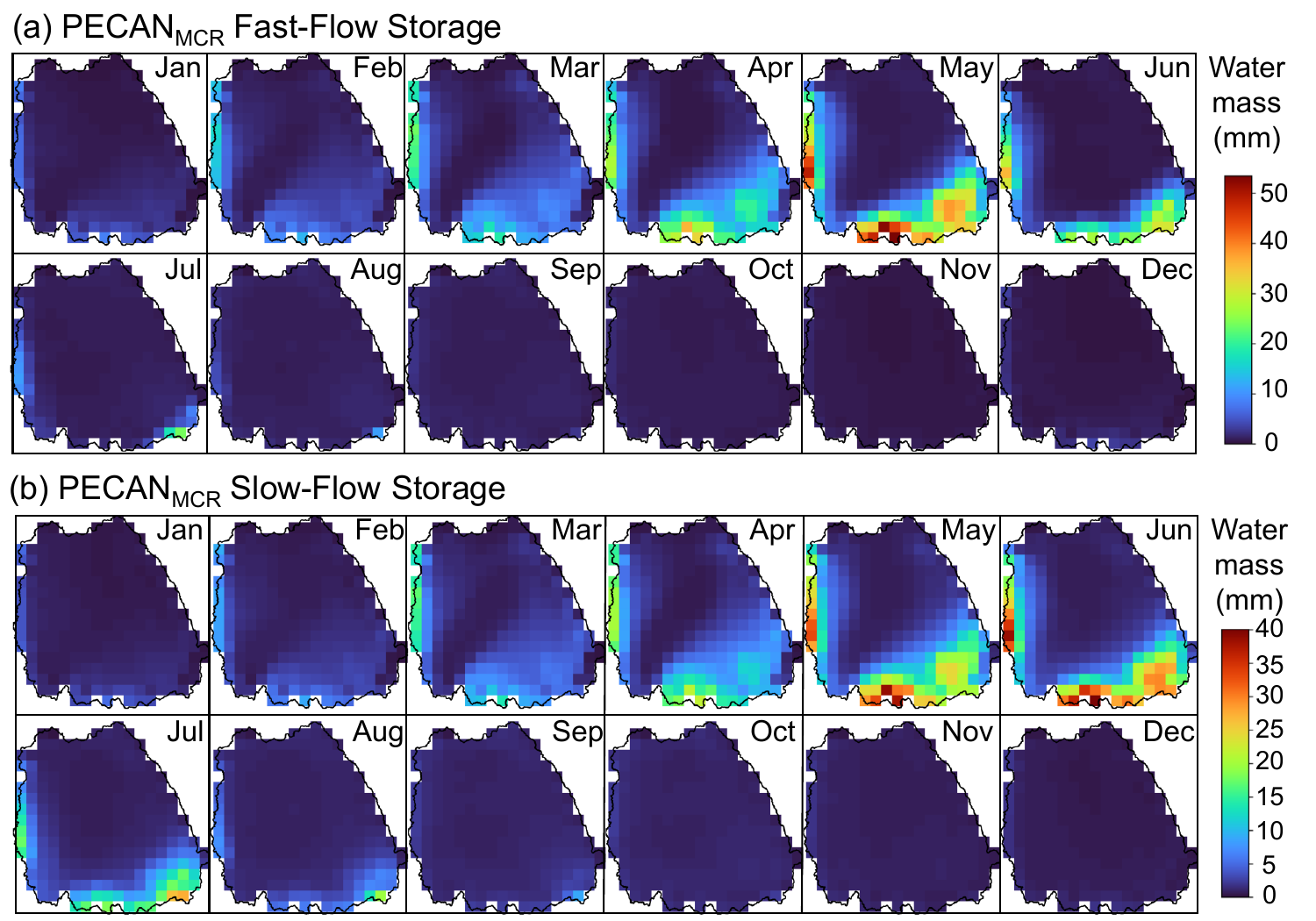}
\vspace{-0.7cm}
\caption{Spatial monthly climatology of (a) PECAN\textsubscript{MCR} fast-flow storage and (b) PECAN\textsubscript{MCR} slow-flow storage.}
\label{fig:S2}
\end{figure}

\begin{figure}[p]
\centering
\includegraphics[width=0.88\textwidth]{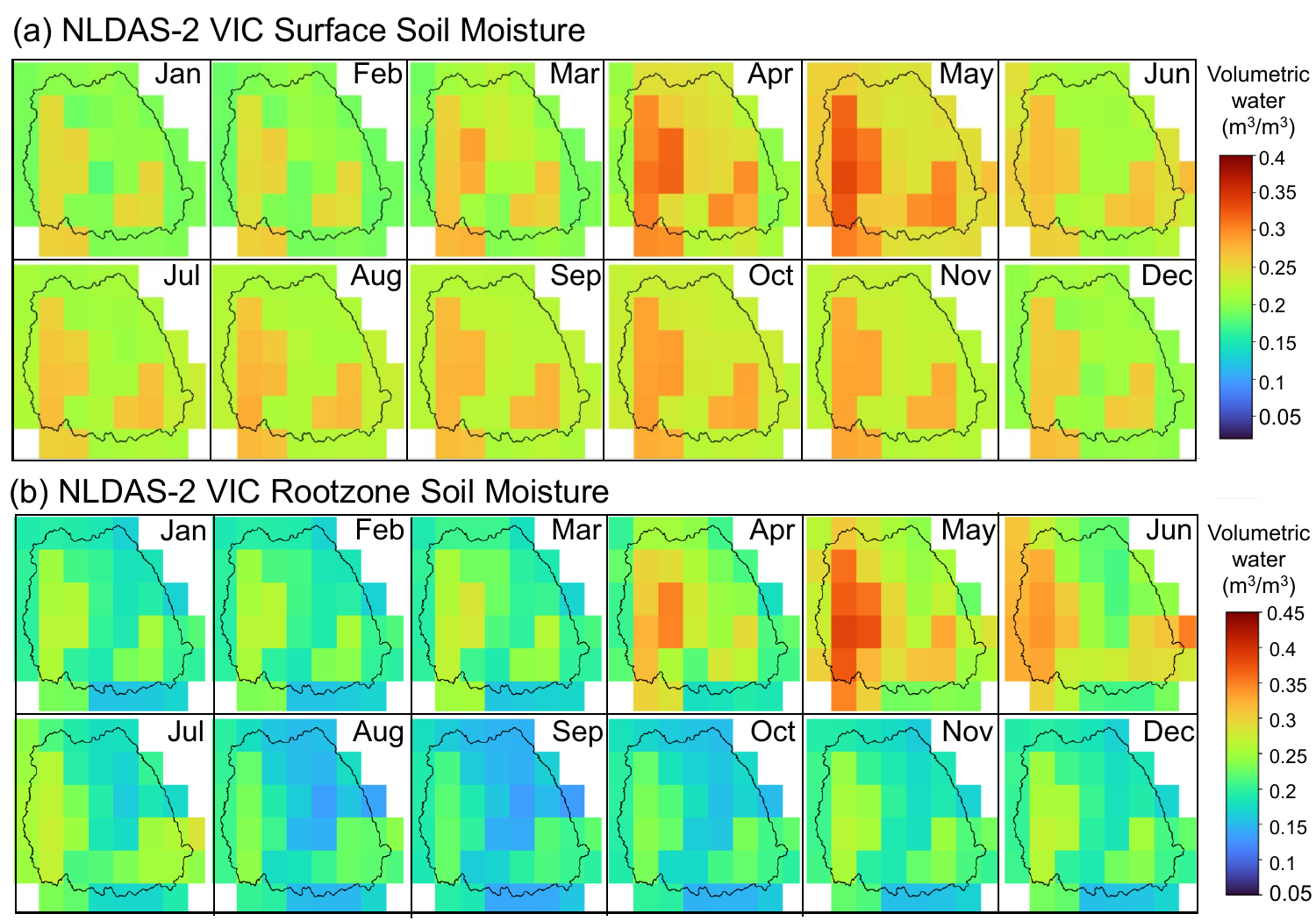}
\vspace{-0.5cm}
\caption{Spatial monthly climatology of NLDAS-2 VIC (a) surface soil moisture (0--10 cm) and (b) root zone soil moisture (0--100 cm) over the study watershed during the validation period (WY2013--WY2022). Obtained from the NLDAS VIC Land Surface Model L4 Monthly 0.125$^\circ$ $\times$ 0.125$^\circ$ V2.0 \citep{xia2012continental}, NASA Goddard Earth Sciences Data and Information Services Center (GES DISC).}
\label{fig:S3}
\vspace{0.8cm}
\includegraphics[width=0.88\textwidth]{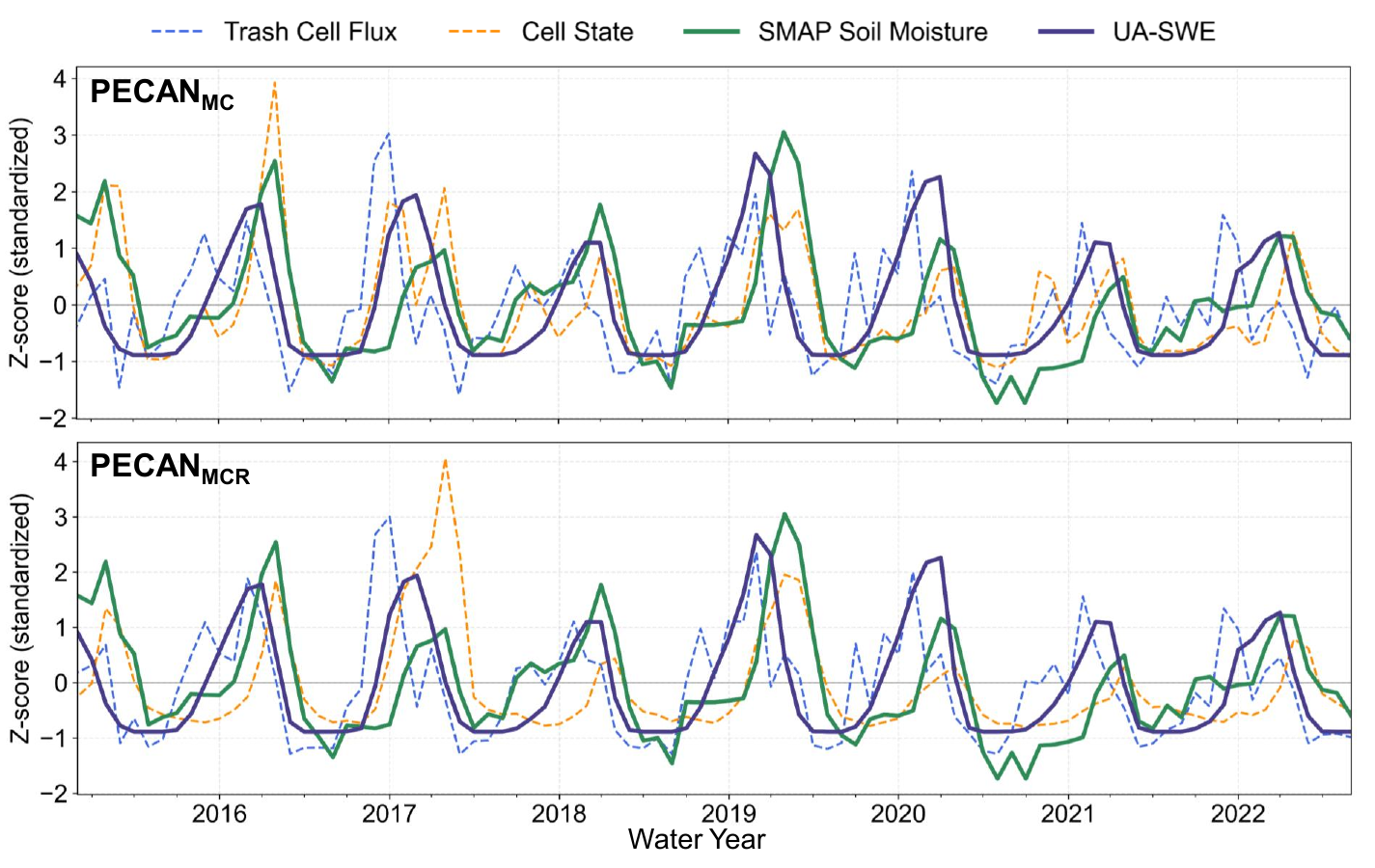}
\vspace{-0.5cm}
\caption{Monthly mean time series of PECAN\textsubscript{MC} (top) and PECAN\textsubscript{MCR} (bottom) cell state and trash cell flux, compared with UA-SWE and SMAP soil moisture averaged across the surface and root-zone layers. All variables are standardized using z-scores (zero mean and unit variance) to facilitate comparison across different physical units and magnitudes. The analysis is restricted to 2015--2022 based on SMAP data availability.}
\label{fig:S4}
\end{figure}
\clearpage

\begin{table}[!htbp]
\centering
\caption{Static catchment attributes for regional LSTM, MC-LSTM and MCR-LSTM training. Variable descriptions are adopted from \citet{kratzert2019toward}.}
\label{tab:S1}
\vspace{+0.1cm}
\small
\begin{tabular}{c p{4.2cm} p{9.5cm}}
\toprule
\textbf{No.} & \textbf{Variable} & \textbf{Description} \\
\midrule
1  & Precipitation mean          & Mean daily precipitation. \\
2  & PET mean                    & Mean daily potential evapotranspiration \\
3  & Aridity index               & Ratio of Mean PET to Mean Precipitation \\
4  & Precip seasonality          & Estimated by representing annual precipitation and temperature as sin waves. Positive (negative) values indicate precipitation peaks during the summer (winter). Values of $\sim$0 indicate uniform precipitation throughout the year. \\
5  & Snow fraction               & Fraction of precipitation falling on days with temp $<$ 0 $^\circ$C. \\
6  & High precipitation frequency & Frequency of days with $\leq$5$\times$ mean daily precipitation \\
7  & High precip duration        & Average duration of high precipitation events (number of consecutive days with $\leq$5$\times$ mean daily precipitation). \\
8  & Low precip frequency        & Frequency of dry days ($<$ 1 mm/day). \\
9  & Low precip duration         & Average duration of dry periods (number of consecutive days with precipitation $<$ 1 mm/day). \\
10 & Elevation                   & Catchment mean elevation. \\
11 & Slope                       & Catchment mean slope. \\
12 & Area                        & Catchment area. \\
13 & Forest fraction             & Fraction of catchment covered by forest. \\
14 & LAI max                     & Maximum monthly mean of leaf area index. \\
15 & LAI difference              & Difference between the max. and min. mean of the leaf area index. \\
16 & GVF max                     & Maximum monthly mean of green vegetation fraction. \\
17 & GVF difference              & Difference between the maximum and minimum monthly mean of the green vegetation fraction. \\
18 & Soil depth (Pelletier)      & Depth to bedrock (maximum 50~m). \\
19 & Soil depth (STATSGO)        & Soil depth (maximum 1.5~m). \\
20 & Soil Porosity               & Volumetric porosity. \\
21 & Soil conductivity           & Saturated hydraulic conductivity. \\
22 & Max water content           & Maximum water content of the soil. \\
23 & Sand fraction               & Fraction of sand in the soil. \\
24 & Silt fraction               & Fraction of silt in the soil. \\
25 & Clay fraction               & Fraction of clay in the soil. \\
26 & Carbonate rocks fraction    & Fraction of the catchment area characterized as ``carbonate sedimentary rocks''. \\
27 & Geological permeability     & Surface permeability (log10). \\
\bottomrule
\end{tabular}
\end{table}

\end{document}